\documentclass[conference]{IEEEtran}
\IEEEoverridecommandlockouts
\usepackage{cite}
\usepackage{amsmath,amssymb,amsfonts}
\usepackage{url}
\usepackage{multirow}
\usepackage{algorithm}
\usepackage[export]{adjustbox}
\usepackage{algpseudocode}
\usepackage{graphicx}
\usepackage{textcomp}
\usepackage{xcolor}
\def\BibTeX{{\rm B\kern-.05em{\sc i\kern-.025em b}\kern-.08em
    T\kern-.1667em\lower.7ex\hbox{E}\kern-.125emX}}
\begin{document}

%\title{A Study of Cluster-Based Cross-Validation Strategies}
\title{Comparing Cluster-Based Cross-Validation Strategies for Machine Learning Model Evaluation}
\author{\IEEEauthorblockN{Afonso Martini Spezia}
\IEEEauthorblockA{\textit{Instituto de Informática} \\
\textit{Universidade Federal do Rio Grande do Sul (UFRGS)} \\
Porto Alegre, Brazil \\
amspezia@inf.ufrgs.br \\ ORCID: 0009-0007-6404-7181}
\and
\IEEEauthorblockN{Thomas Fontanari}
\IEEEauthorblockA{\textit{Instituto de Informática} \\
\textit{Universidade Federal do Rio Grande do Sul (UFRGS)} \\
Porto Alegre, Brazil \\
thomas.fontanari@inf.ufrgs.br}
\and
\IEEEauthorblockN{Mariana Recamonde-Mendoza}
\IEEEauthorblockA{\textit{Instituto de Informática} \\
\textit{Universidade Federal do Rio Grande do Sul (UFRGS)} \\
Porto Alegre, Brazil \\
mrmendoza@inf.ufrgs.br \\ ORCID: 0000-0003-2800-1032}
}

\maketitle

\begin{abstract}
Cross-validation plays a fundamental role in Machine Learning, enabling robust evaluation of model performance and preventing overestimation on training and validation data. However, one of its drawbacks is the potential to create data subsets (folds) that do not adequately represent the diversity of the original dataset, which can lead to biased performance estimates. The objective of this work is to deepen the investigation of cluster-based cross-validation strategies by analyzing the performance of different clustering algorithms through experimental comparison. Additionally, a new cross-validation technique that combines Mini Batch K-Means with class stratification is proposed. Experiments were conducted on 20 datasets (both balanced and imbalanced) using four supervised learning algorithms, comparing cross-validation strategies in terms of bias, variance, and computational cost. The technique that uses Mini Batch K-Means with class stratification outperformed others in terms of bias and variance on balanced datasets, though it did not significantly reduce computational cost. On imbalanced datasets, traditional stratified cross-validation consistently performed better, showing lower bias, variance, and computational cost, making it a safe choice for performance evaluation in scenarios with class imbalance. In the comparison of different clustering algorithms, no single algorithm consistently stood out as superior. Overall, this work contributes to improving predictive model evaluation strategies by providing a deeper understanding of the potential of cluster-based data splitting techniques and reaffirming the effectiveness of well-established strategies like stratified cross-validation. Moreover, it highlights perspectives for increasing the robustness and reliability of model evaluations, especially in datasets with clustering characteristics.
\end{abstract}


\section{Introduction}
The field of Machine Learning (ML) has gained exponential relevance in recent years due to its applications across various domains, including healthcare, business, industry, and the military \cite{Aggarwal}. The models developed enable systems to learn from large datasets, allowing the detection of complex patterns and decision-making based on them. However, the reliability of these models’ results depends on the quality of the metrics used to evaluate them. Obtaining accurate metrics is fundamental to understanding the true performance of these models, thus highlighting the need for robust evaluation tools.

Cross-validation stands out in this context as a technique for the evaluation and selection of predictive models. As the most widely used evaluation method today, it works by splitting the dataset into parts (called folds) that are used to train and test the models iteratively. This helps ensure that models are robust and generalize well to unseen data, avoiding issues such as performance overestimation.

In recent years, several studies have proposed more accurate and efficient ways to evaluate machine learning models. Through class stratification and cluster-based splitting, for instance, more precise assessments can be achieved. However, some of these solutions, such as cluster-based data splits, can be computationally expensive on large datasets \cite{Ikotun}.

In a previous study, Fontanari et al.\cite{Fontanari} proposed the use of Mini-Batch K-Means to reduce the processing time required for clustering. Nevertheless, the authors showed that although more elaborate cross-validation strategies may offer potential gains in model evaluation, stratified cross-validation remains preferable when using a small number of folds or in imbalanced datasets.

In this regard, the present work proposes an implementation of K-Fold and Mini-Batch K-Fold with class stratification for cluster-based data splitting, aiming to capture possible intraclass subgroups that might not be detected by other techniques. The goal is to achieve estimates with balanced bias, variance, and computational cost. Additionally, this work conducts a comparative study between well-established data splitting methods and the proposed technique, as well as among different clustering algorithms, namely K-Means, DBSCAN \cite{sander1998gdbscan}, and Agglomerative Hierarchical Clustering \cite{Diamantidis}. During the experiments, different data splitting strategies are analyzed to assess whether any method stands out in terms of bias, variance, and computational cost.

The remainder of this paper is structured as follows. Section~\ref{sec:theoretical_background} presents the theoretical background necessary to understand the concepts of machine learning, cross-validation, and the algorithms used in this work. Section~\ref{cap:revisaoBibliografica} discusses related studies that address cross-validation techniques and their impact on model evaluation, comparing them with the proposal of this study. Section~\ref{sec:methodology} details the methodology adopted, including the implementation of stratified Mini-Batch K-Fold and the clustering algorithms used. Section~\ref{sec:results} presents the experimental results and the comparative analysis between the methods, focusing on bias, variance, and computational cost. Finally, Section~\ref{sec:conclusion} offers the conclusions based on the results obtained and outlines possible directions for future work.a

%This study is structured into six chapters. Chapter 2 presents the theoretical background necessary to understand the concepts of machine learning, cross-validation, and the algorithms used in this work. Chapter 3 discusses related studies that address cross-validation techniques and their impact on model evaluation, comparing them with the proposal of this study. Chapter 4 details the methodology adopted, including the implementation of stratified Mini-Batch K-Fold and the clustering algorithms used. Chapter 5 presents the experimental results and the comparative analysis between the methods, focusing on bias, variance, and computational cost. Finally, Chapter 6 offers the conclusions based on the results obtained and outlines possible directions for future work.

\section{Theoretical Background}
\label{sec:theoretical_background}

This section covers the key concepts related to the development of this work, including a brief review of supervised and unsupervised machine learning, as well as selected algorithms for each learning type. The validation techniques used to evaluate the predictive models generated through machine learning are also discussed, with a particular focus on cross-validation and its variations.

\subsection{Machine Learning}

Machine Learning (ML) refers to a broad range of algorithms used to make predictions and detect patterns automatically in datasets. Recent advancements in the field have achieved levels of semantic understanding and information extraction comparable to, or even surpassing, human capabilities. Its exponential growth in recent decades is due to the increasing volume of data, rapid development of computational resources, and progress in algorithm design \cite{Nichols}.

In general, ML describes the ability of systems to learn and make inferences from datasets related to specific problems. This enables the automation of analytical model construction and the solving of associated tasks \cite{bishop2006pattern}. Among the main types of learning, supervised and unsupervised learning are particularly relevant and are reviewed in the following subsections.

\subsubsection{Supervised Learning}

Supervised learning is an ML technique in which a model is trained using a labeled dataset. Each data point in the training set consists of an input and the corresponding desired output, or label. The goal of supervised learning algorithms is to learn to map inputs to correct outputs so that the model can make accurate predictions on unseen data.

This type of learning has been widely studied and applied across multiple domains, proving effective in solving complex classification and regression problems. The review by Mohri\cite{Mohri} highlights the broad applicability and effectiveness of these techniques in areas such as speech recognition, computer vision, bioinformatics, and fraud detection.

In the context of this work, supervised learning techniques such as stratified cross-validation are used to assess the performance of predictive models on labeled data. Additionally, supervised algorithms are used to train models that aim to establish a relationship between input data and output labels, referred to in this work as classes.

\subsubsection{Logistic Regression}

Logistic Regression (LR) is a widely used machine learning algorithm for classification problems, especially when the goal is to predict the probability of occurrence of a binary class. As discussed by Menard \cite{menard2002}, LR is robust in situations where the relationships between independent variables and the dependent variable are not perfectly linear, which broadens its applicability to real-world data, where such relationships are often more complex.

It works by modeling the relationship between one or more independent variables and a binary dependent variable. The binary dependent variable can take two possible values, usually coded as 0 and 1, representing, for instance, a negative or positive class. First, a linear combination of these variables is created and then passed through the sigmoid function to convert it into a probability between 0 and 1. Finally, this probability is used to make a classification, typically by comparing it to a threshold, such as 0.5. If the probability is greater than the threshold, the model predicts one class; otherwise, it predicts the other.

\subsubsection{Decision Tree}

Decision Tree (DT) is another machine learning algorithm used for both classification and regression problems. The algorithm builds a tree-shaped model from a dataset, where each internal node represents a condition on an input variable, each branch represents the outcome of that condition, and each leaf node represents a class prediction, in the case of classification problems.

\begin{figure}[htbp]
\centerline{\includegraphics[width=\linewidth]{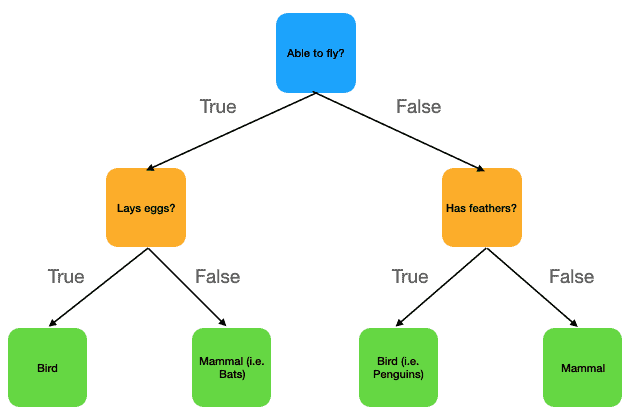}}
\caption{Example of a Decision Tree classifying animals according to their characteristics.}
\label{fig:dt}
\end{figure}

According to Breiman et al.\cite{breiman1984}, DTs are easily interpretable and visualizable, which facilitates the understanding of the patterns and decisions made by the model. Figure \ref{fig:dt} illustrates the decision process in the algorithm, where each node asks a specific question, and the path followed through the branches leads to the final classification of the problem.

\subsubsection{Support Vector Machines}

Support Vector Machines (SVM) are machine learning algorithms primarily used for classification problems. They work by finding the hyperplane that maximizes the separation between classes. When the classes are not linearly separable, the SVM uses the "kernel trick" to map the data into a higher-dimensional space, where separation becomes possible. According to Cortes and Vapnik \cite{cortes1995}, this approach enables SVMs to effectively handle complex and high-dimensional data.

Moreover, SVMs are known for their robustness and ability to avoid overfitting, especially when the classes are clearly separable. They can also be adapted to different types of problems through the selection of kernel functions, such as the Radial Basis Function (RBF) and polynomial kernels. This flexibility makes SVMs a popular choice across a range of applications, from bioinformatics to text processing.

\subsubsection{Random Forest}

Random Forest (RF) is a machine learning algorithm used for both classification and regression. It operates by creating a set of independent Decision Trees from different subsets of the original dataset and combining their predictions to improve accuracy and reduce the risk of overfitting. Breiman et al.\cite{breiman2001} introduced the concept of RF, highlighting its ability to handle large datasets with many features, which makes the algorithm highly effective in various contexts.

\subsection{Unsupervised Learning}

Unlike supervised learning, unsupervised learning works with unlabeled data. Its goal is to discover hidden patterns or structures in the data without relying on predefined labels. A common task in unsupervised learning is data clustering, using algorithms such as K-Means \cite{MacQueen1967} and DBSCAN \cite{Ester1996}. This type of learning is widely applied in tasks like market segmentation, data compression, and dimensionality reduction.

In this work, clustering algorithms are used to create clusters from the data, which are then employed in cross-validation to improve the representativeness of the folds. Three such algorithms are explored: K-Means (including its Mini-Batch variant), DBSCAN, and Agglomerative Clustering. Each has distinct characteristics and specific applications, making them suitable for different types of data and problems. The evaluation in this work favors algorithms that perform well in general, without focusing on specific dataset characteristics. Thus, it is expected that the performance of some algorithms will stand out among others in the selected datasets, providing a broad and comparative view of the effectiveness of each method.

\subsubsection{K-Means}

K-Means is one of the simplest and most widely used clustering algorithms. It partitions a dataset into \textit{K} clusters, where each data point belongs to the cluster with the nearest centroid. The algorithm follows these steps:

\begin{enumerate}
    \item \textbf{Initial centroid definition:} Define the initial centers of the \( K \) clusters, where \( K \) is a user-defined parameter. The initial centroids may be randomly assigned or pre-defined using other algorithms.
    \item \textbf{Distance calculation and cluster assignment:} Compute the distance from each point to each centroid, and assign each data point to the cluster whose centroid is closest.
    \item \textbf{Centroid recomputation:} Recalculate the centroids based on the elements within each cluster.
\end{enumerate}

Steps 2 and 3 are then repeated until there are no significant changes in cluster assignments between iterations.

Although K-Means is efficient for moderately sized datasets, its efficiency significantly decreases as the dataset size grows. This is due to the need to compute distances from each point to all centroids in every iteration, resulting in high computational cost for large-scale data. Additionally, K-Means is sensitive to centroid initialization, often requiring multiple runs to achieve good results, which can further increase processing time.

\subsubsection{Mini Batch K-Means}

Mini Batch K-Means is a variation of the K-Means algorithm designed to improve computational efficiency when handling large datasets. Instead of using the entire dataset at each iteration, Mini Batch K-Means processes small random subsets of data called mini-batches. In each iteration, a mini-batch is randomly selected from the full dataset, and each point in the mini-batch is assigned to the nearest cluster center. The cluster centers are then updated based on the points in the mini-batch, and this process is repeated with new mini-batches until the cluster centers stabilize or a maximum number of iterations is reached.

The main advantages of Mini Batch K-Means include improved speed and scalability. By processing only a small fraction of the data in each iteration, it significantly reduces processing time compared to traditional K-Means. Although it may be slightly less accurate, Mini Batch K-Means generally converges faster, achieving a balance between precision and efficiency \cite{Sculley2010}.

\subsubsection{DBSCAN}

DBSCAN (Density-Based Spatial Clustering of Applications with Noise) is a density-based clustering algorithm capable of identifying clusters with complex shapes and detecting noise points in a dataset. It takes as input the parameters $\varepsilon$, the search radius around each point, and \textit{min\_samples}, the minimum number of points required to form a cluster, and operates as follows:

\begin{itemize}
    \item \textbf{Neighbor identification:} For each point, identify neighboring points within the $\varepsilon$ radius;
    \item \textbf{Point classification:} Classify each point as a core point (with at least \textit{min\_samples} neighbors), a border point (not a core point but within the neighborhood of a core point), or a noise point (not a core point and lacking sufficient neighbors);
    \item \textbf{Cluster formation:} Create clusters by connecting core points and their neighbors.
\end{itemize}

DBSCAN is effective in forming clusters with irregular shapes and is robust to noise. However, the correct choice of the parameters $\varepsilon$ and \textit{min\_samples} is crucial for its performance. A study by Schubert et al. \cite{schubert2017dbscan} showed that DBSCAN's effectiveness is highly dependent on parameter selection, especially in datasets with varying densities. In such cases, the sensitivity of $\varepsilon$ may lead to false cluster detection or failure to identify true clusters. The authors also proposed improvements to the algorithm that adapt $\varepsilon$ dynamically, resulting in better cluster detection when density is non-uniform.

\subsubsection{Agglomerative Clustering}

Agglomerative Clustering is a hierarchical algorithm that builds a tree of clusters using a bottom-up approach. This method is known for its simplicity and effectiveness in creating hierarchical clusters, which can be visualized as dendrograms. The process is described as follows:

\begin{itemize}
    \item \textbf{Initialization:} Each data point is initially considered as its own cluster.
    \item \textbf{Cluster merging:} Iteratively, the two closest clusters are merged until all points belong to a single cluster or a stopping criterion is reached.
\end{itemize}

There are different methods to define the proximity between clusters, such as single linkage, complete linkage, and average linkage. The algorithm is flexible, allowing analysis at different levels of granularity, but it can be computationally intensive for large datasets \cite{Murtagh2014}.

\subsection{Cross-Validation}

Cross-validation is a technique used to robustly evaluate the performance of a predictive model and to reduce the risk of overfitting, which occurs when the model fits too closely to the training data. This leads to an overestimation of its performance and, consequently, to poor generalization on test data (i.e., unseen data) \cite{maleki2020machine}. The technique involves training multiple machine learning models on subsets of the available input data and evaluating them on the complementary subset. There are various variations of this technique, depending on how the data subsets are generated. Some of these variations will be discussed below.

\subsection{K-Fold Cross-Validation}

\begin{figure}[htbp]
\centering
\includegraphics[width=\linewidth]{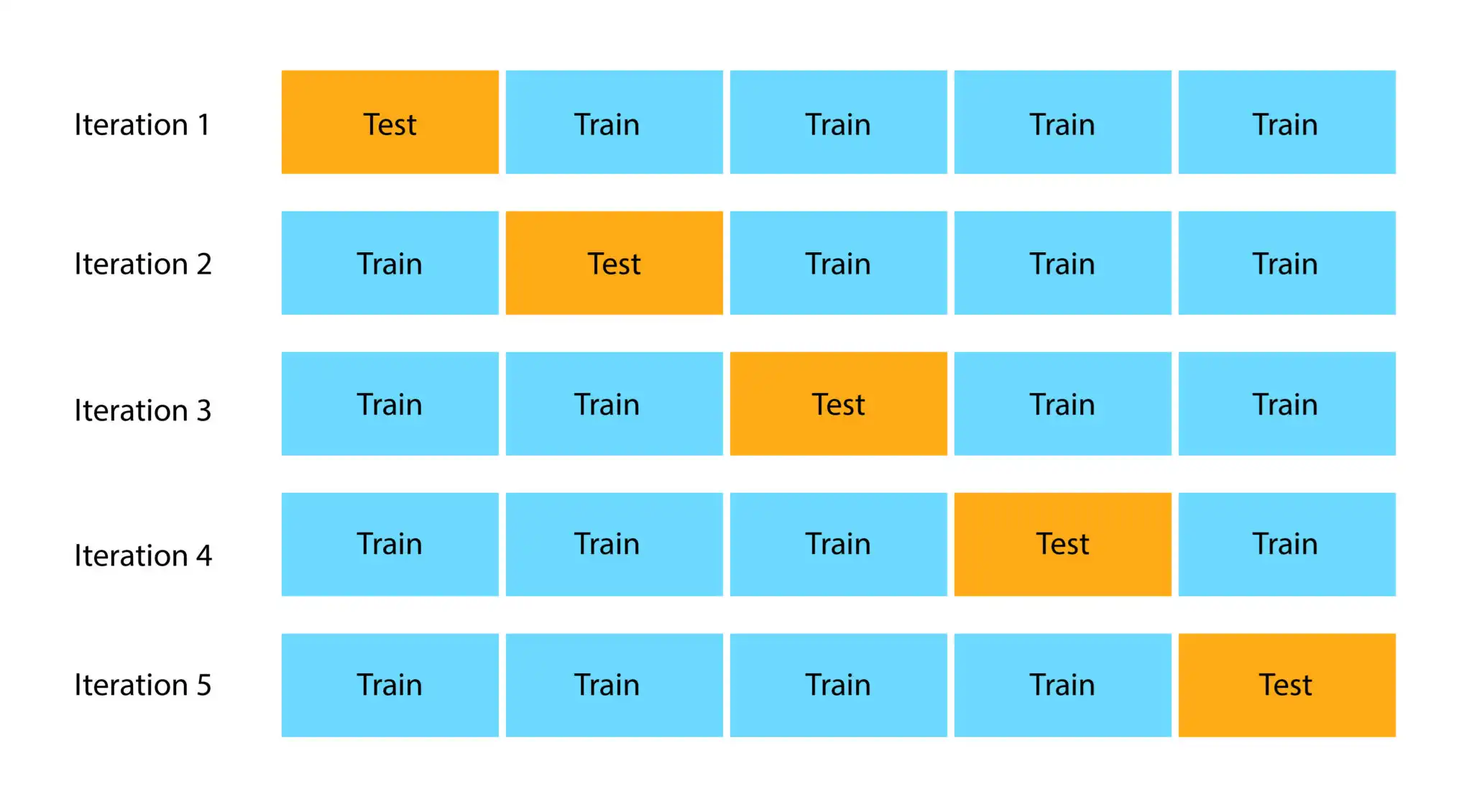}
\caption{K-Fold Cross-Validation using 5 folds.}
\label{fig:validacaocruzadakfold}
\end{figure}

Among cross-validation techniques, the most commonly used is K-Fold Cross-Validation. In this method, the dataset is divided into \( K \) parts, called folds. Then, the model is trained on \( K-1 \) parts and tested on the remaining part. This process is repeated \( K \) times, with a different fold being used for validation in each iteration. Finally, the model’s performance is computed as the arithmetic mean of the performance across all iterations. Figure \ref{fig:validacaocruzadakfold} illustrates the training and testing process using K-Fold Cross-Validation with five folds.

However, due to the natural randomness of this method when splitting the data, the generated folds may not adequately represent the dataset as a whole. As a result, a given fold might contain a disproportionate number of instances from a specific class compared to the full dataset, while some classes might even be absent from other folds. This variability in fold composition can cause strong instability in the results of K-Fold Cross-Validation, particularly for imbalanced datasets, thus affecting the reliability of model performance metrics. Therefore, it is crucial to consider additional strategies to mitigate the effects of this variability and ensure a more consistent representation of the dataset during model evaluation.

\subsection{Stratified Cross-Validation}

Stratified Cross-Validation (SCV) is especially useful when dealing with imbalanced datasets, where class distributions are highly uneven. This technique ensures that each fold maintains the original proportion of instances from each class. Preserving this class distribution is crucial for preventing bias in model evaluation, as it guarantees that all classes are represented in both the training and test sets during each iteration of the validation process.

\subsection{Cluster-Based Cross-Validation}

\begin{figure}[htbp]
\centering
\includegraphics[width=\linewidth]{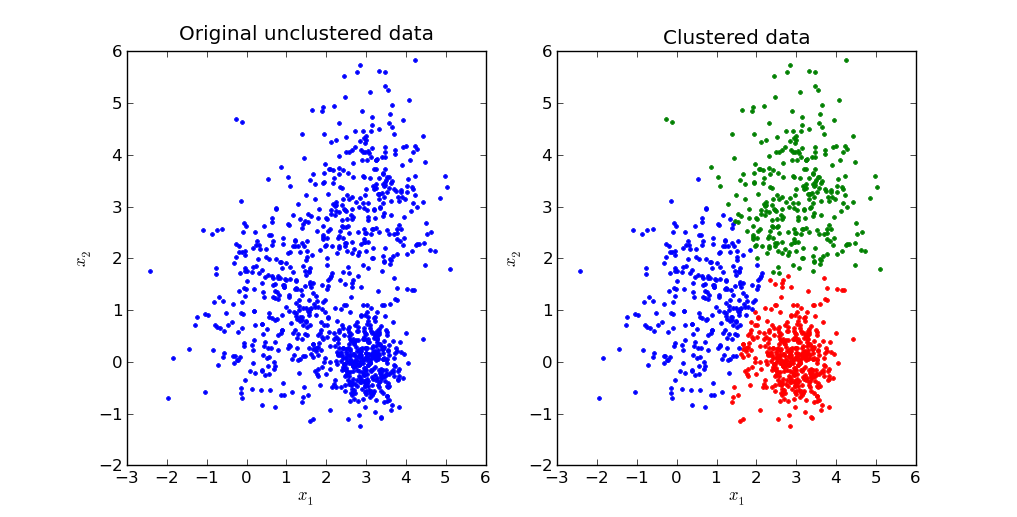}
\caption{Example of applying a clustering algorithm to a dataset.}
\label{fig:clusters}
\end{figure}

Clustering, that is, grouping similar data points, is a central technique in unsupervised learning, where the true class labels of the data are unknown. By grouping data into clusters, clustering algorithms help identify and separate different subgroups within the dataset. This facilitates data analysis and interpretation, revealing patterns that may not be immediately apparent and enabling a better understanding of the relationships among the data variables. Figure \ref{fig:clusters} illustrates an example of how a clustering algorithm such as K-Means can segment a dataset into multiple clusters by grouping together similar instances.

\vspace{0.5cm}

In the context of cross-validation, splitting data based on clusters has been proposed to create folds that are more representative of the original dataset \cite{Diamantidis}. Clusters are formed using clustering algorithms such as K-Means, which group similar data points into distinct clusters. Then, data points from the same cluster are distributed across different folds in the cross-validation process. This approach aims to ensure that each fold contains a more faithful representation of the entire dataset, even when the data is imbalanced. Some cluster-based cross-validation methods will be discussed in more detail in Section~\ref{cap:revisaoBibliografica}.

\section{Related Work}
\label{cap:revisaoBibliografica}

Cross-validation and clustering algorithms have been extensively studied in the literature due to their importance in developing and evaluating machine learning models. Various methods have been proposed to improve the efficiency and accuracy of these processes, especially in contexts involving large-scale and imbalanced data.

This section explores works that, like the present study, aim to enhance cross-validation and clustering techniques. The review covers studies using both supervised and unsupervised learning and introduces novel cross-validation methods, such as cluster-based validation, as well as adaptations that improve scalability and accuracy, such as Mini Batch K-Means. Furthermore, it discusses advances and comparisons among different clustering algorithms, highlighting their practical applications and contributions to better data analysis in machine learning.

\subsection{Distribution-balanced Stratified Cross-Validation (DBSCV)}

Zeng et al.\cite{Zeng} proposed the \textit{Distribution-balanced Stratified Cross-Validation} (DBSCV) technique to address limitations in traditional cross-validation with imbalanced datasets, aiming to ensure that each fold preserves the original class distribution. This method seeks to generate folds that are representative of the entire dataset by assigning neighboring instances to different folds.

The process starts by randomly selecting an instance, which is then assigned to a fold. Next, the algorithm moves to the nearest instance of the same class and assigns it to the next fold, repeating this procedure until all instances of that class have been distributed. The same procedure is applied to the remaining classes, ensuring each fold contains approximately the same number of instances per class.

\subsection{Distribution Optimally Stratified Cross-Validation (DOBSCV)}

Moreno et al.\cite{Moreno} proposed the \textit{Distribution Optimally Balanced Stratified Cross-Validation} (DOBSCV) technique as an improvement over DBSCV, initially introduced by Zeng \cite{Zeng}. The goal of DOBSCV is to further reduce bias and variation in cross-validation by ensuring that the folds are not only class-balanced but also optimized to reduce interference between training and testing distributions. DOBSCV seeks to ensure that each fold is a faithful representation of the original dataset, mitigating the effects of dataset shift—a condition where training and testing data distributions differ and negatively impact model validation.

DOBSCV works similarly to DBSCV, starting with a random instance from the dataset. However, instead of finding the nearest instance of the same class, it identifies the \(k - 1\) nearest neighbors (where \(k\) is the number of folds) of the current instance from the same class and assigns each one to a different fold. This process is repeated independently for each class, similarly to DBSCV, until all instances have been assigned.

\subsection{Unsupervised Stratification of Cross-Validation for Accuracy Estimation}

Diamantidis et al.\cite{Diamantidis} introduced an innovative approach to cross-validation to address the problem of underrepresentation of classes in randomly generated folds. In their paper, the authors proposed using clustering techniques to improve fold representativeness in cross-validation. The main idea is to divide the dataset into clusters using algorithms such as Hierarchical Agglomerative Clustering and K-Means.

The proposed method begins by applying a clustering algorithm, such as K-Means or Agglomerative Clustering, to divide the dataset into \(m\) clusters. Once the clusters are formed, their elements are ordered based on their distance to the cluster center. These elements are then distributed across the \(k\) folds of the cross-validation process, ensuring that adjacent instances are assigned to different folds. This approach ensures that each fold contains a more diverse and balanced representation of the original dataset, reducing bias and yielding more accurate performance estimates of machine learning models.

\subsection{Cross-validation Strategies for Balanced and Imbalanced Datasets}

Fontanari, Fróes, and Recamonde-Mendoza \cite{Fontanari} proposed an adaptation of the cluster-based cross-validation algorithm using Mini-Batch K-Means, with the objective of reducing computational cost in comparison to traditional K-Means. Originally, the K-Means algorithm can be computationally intensive, especially in scenarios involving large datasets, due to the need to recalculate the centroids of all data points at each iteration. To address this issue, the article explores the use of Mini-Batch K-Means, which processes only a small fraction of the data at each iteration, significantly reducing execution time and computational load.

The authors implemented and compared several data partitioning strategies, including Distribution-Based Stratified Cross-Validation (DBSCV), Distribution-Optimized Balanced Stratified Cross-Validation (DOBSCV), and Cluster-Based Distribution Stratified Cross-Validation (CBDSCV). The main highlight of the study was the adaptation of CBDSCV using Mini-Batch K-Means, which enabled a significant reduction in execution time while maintaining performance comparable to the traditional CBDSCV method. The Mini-Batch K-Means approach selects only a sample of the data at each iteration, thereby decreasing computational cost without significantly compromising the quality of the obtained results.

The experiments conducted by the authors evaluated the effectiveness of the different strategies across datasets of varying sizes and class imbalance levels. The results indicated that the more elaborate cross-validation strategies showed potential gains in scenarios with a small number of folds, but stratified cross-validation was preferable in scenarios with 10 folds or in imbalanced datasets. The adaptation of CBDSCV with Mini-Batch K-Means proved effective in reducing computational cost, becoming a viable alternative in situations where efficiency is crucial.

\subsection{Summary}

In this section, several cross-validation techniques were reviewed, focusing on methods that seek to improve representativeness and efficiency in imbalanced-data scenarios. Distribution-balanced Stratified Cross-Validation (DBSCV) and its optimization, Distribution Optimally Balanced Stratified Cross-Validation (DOBSCV), proved effective in addressing the limitations of traditional cross-validation by ensuring that folds preserve the original class distribution. In addition, techniques incorporating clustering methods, such as those proposed by Diamantidis et al. \cite{Diamantidis}, demonstrated the potential to improve data distribution among folds, reducing bias and providing more accurate evaluations. However, these approaches also introduce new challenges, such as computational complexity and scalability, especially when applied to large data volumes.

Despite these advances, important gaps remain in the literature, particularly regarding the balance between data representativeness and computational efficiency. The work of Fontanari et al. \cite{Fontanari} with Mini-Batch K-Means offers a promising solution to reduce computational cost, yet further adaptations and combinations of techniques are needed to optimize these processes even more. The present study will focus on exploring and developing new strategies that integrate more advanced clustering methods with cross-validation and class stratification, aiming not only to enhance the performance and robustness of cross-validation procedures but also to ensure that these techniques are scalable and applicable across different domains.

\section{Methodology}
\label{sec:methodology}

The main objective of this work is to evaluate various cross-validation techniques, with an emphasis on comparing traditional methods and those based on clustering algorithms. This study proposes the implementation of a new technique that combines K-Means and Mini-Batch K-Means with class stratification, aiming to strike a balance between performance and computational efficiency. This section presents the methodology used in the conducted experiments. Below, the cross-validation methods employed in the experiments are detailed and named according to their specific approach:

\textbf{SCBCV} (Stratified Cluster-Based Cross-Validation): A cross-validation technique proposed in this work, which combines class stratification with the K-Means clustering algorithm;

\textbf{SCBCV Mini}: A version of SCBCV that employs Mini-Batch K-Means;

\textbf{KCBCV} (K-Means Cluster-Based Cross-Validation): A cross-validation technique that uses the K-Means algorithm for cluster formation, without class stratification;

\textbf{KCBCV Mini}: A version of KCBCV that uses Mini-Batch K-Means;

\textbf{ACBCV} (Agglomerative Cluster-Based Cross-Validation): A technique that applies the Agglomerative Clustering algorithm for cluster formation;

\textbf{DBSCANBCV} (DBSCAN-Based Cross-Validation): A technique that employs the DBSCAN clustering algorithm for cluster formation;

\textbf{SCV} (Stratified Cross-Validation): A k-fold cross-validation technique stratified by class, well-established in the literature.

Finally, the experiments proposed in this study were divided into three distinct sets, described below:

\textbf{1st Set}: Comparison between SCBCV using a variable number of clusters defined for each dataset and fixed values of 2, 3, 4, and 5 clusters;

\textbf{2nd Set}: Comparison among SCBCV, SCBCV Mini, and SCV;

\textbf{3rd Set}: Comparison among the different types of cluster-based cross-validation techniques discussed in this work — SCBCV, SCBCV Mini, KCBCV, KCBCV Mini, DBSCANBCV, and ACBCV.

\subsection{Datasets}

The selection of datasets for this study was based on their diverse and comprehensive characteristics among those available in PMLB (Penn Machine Learning Benchmarks) \cite{Olson2017PMLB}. PMLB is an extensive collection of standardized datasets designed for the evaluation and comparison of machine learning algorithms. The chosen datasets vary in terms of number of classes, number of instances, number of attributes, and class imbalance, allowing for a robust analysis of the cross-validation techniques previously described and enabling an assessment of their behavior across different types of datasets.

Table \ref{tbl:datasets} presents the list of datasets used in the experiments, divided into balanced and imbalanced categories. This division was made to analyze the effectiveness of cross-validation techniques in distinct class distribution scenarios, using appropriate metrics for each type. The imbalance of the PMLB datasets is measured by computing the sum of the squared distances between the proportion of instances in each class and the perfectly balanced distribution in the dataset, as shown in Equation \ref{eq:1}, where \( n_i \) is the number of instances of class \( i \), \( K \) is the total number of classes, and  \( N \) is the dataset size.

\begin{equation}
\label{eq:1}
I = K \sum_{i=1}^{K} \left( \frac{n_i}{N} - \frac{1}{K} \right)^2
\end{equation}

The higher the value of index $I$, the greater the class imbalance, with $I$ approaching 1 when almost all instances belong to a single class. In this study, datasets with an imbalance index greater than \(0,20\) were classified as imbalanced \cite{Fontanari}.
As an example, the \textit{haberman} dataset used in the experiment has an imbalance index of \(0.22\) with two classes, where 26.5\% of the data belongs to one class and \(73.5\%\) to the other, characterizing a considerable imbalance. Another dataset used, \textit{dis}, presents an imbalance index of \(0.94\), also with two classes, where \(98.5\%\) of the instances belong to one class and only \(1.5\%\) to the other, illustrating an extreme and significant imbalance.

\begin{table*}[h]
\centering
\caption{List of datasets used in the experiment, divided into balanced and imbalanced distributions and sorted by increasing number of instances.}
\vspace{0.5cm}
\begin{tabular}{|l|c|c|c|c|c|}
\hline
Dataset & Instances & Attributes & Classes & Imbalance & Clusters \\
\hline
\multicolumn{6}{|c|}{Balanced} \\
\hline
cloud & 108 & 7 & 4 & 0,01 & 4 \\
\hline
iris & 150 & 4 & 3 & 0,00 & 4 \\
\hline
analcatdata\_germangss & 400 & 5 & 4 & 0,00 & 4 \\
\hline
movement\_libras & 360 & 90 & 15 & 0,00 & 5 \\
\hline
sonar & 208 & 60 & 2 & 0,00 & 4 \\
\hline
vowel & 990 & 13 & 11 & 0,00 & 4 \\
\hline
contraceptive & 1473 & 9 & 3 & 0,03 & 5 \\
\hline
splice & 3188 & 60 & 3 & 0,08 & 2 \\
\hline
waveform\_21 & 5000 & 21 & 3 & 0,00 & 4 \\
\hline
optdigits & 5620 & 64 & 10 & 0,00 & 5 \\
\hline
\multicolumn{6}{|c|}{Imbalanced} \\
\hline
analcatdata\_cyyoung9302 & 92 & 10 & 2 & 0,34 & 4 \\
\hline
appendicitis & 106 & 7 & 2 & 0,36 & 6 \\
\hline
backache & 180 & 32 & 2 & 0,52 & 5 \\
\hline
new\_thyroid & 215 & 5 & 3 & 0,30 & 6 \\
\hline
haberman & 306 & 3 & 2 & 0,22 & 5 \\
\hline
wine\_quality\_red & 1599 & 11 & 6 & 0,23 & 5 \\
\hline
allrep & 3772 & 29 & 4 & 0,91 & 7 \\
\hline
dis & 3772 & 29 & 2 & 0,94 & 7 \\
\hline
churn & 5000 & 20 & 2 & 0,51 & 4 \\
\hline
ann\_thyroid & 7200 & 21 & 3 & 0,79 & 7 \\
\hline
\end{tabular}
\label{tbl:datasets}
\end{table*}

\subsection{Classification Algorithms and Hyperparameter Optimization}

For the experiments, supervised learning algorithms with varying levels of bias and variance were selected. Specifically, the algorithms presented in Section~\ref{cap:revisaoBibliografica} were used: Logistic Regression (LR), Decision Tree (DT), Support Vector Machines (SVM), and Random Forest (RF).

LR tends to have relatively high bias and low variance, as it assumes a linear relationship between the variables, which may oversimplify the complexity of the data \cite{hastie2009elements}. In contrast, DT exhibits low bias, as it closely fits the training data, but suffers from high variance, which can lead to overfitting, especially when the trees are deep \cite{james2013introduction}. SVM balances bias and variance depending on the choice of kernel; with a linear kernel, bias tends to be higher and variance lower, while more complex kernels, such as RBF, can reduce bias but increase variance \cite{scholkopf2002}. In this study, the default RBF kernel was used, as linear decision functions are already represented by LR. Finally, RF reduces variance without significantly increasing bias by combining multiple DTs \cite{breiman2001}.

Before conducting the experiments, hyperparameter tuning was performed for each learning algorithm on each dataset. Following the approach established in studies such as \cite{hsu2003practical}, Grid Search with 5-fold cross-validation was used to determine the optimal hyperparameters for each algorithm on each dataset. The balanced accuracy metric was employed, as it is more appropriate for handling both balanced and imbalanced datasets. The hyperparameter configurations with the highest scores were selected for the experiments.

The optimized hyperparameters are shown in Table \ref{tbl:hiper}. For LR and SVM, the hyperparameter $C$, which controls model regularization, was tuned, with lower values imposing stronger regularization. For SVM, the $gamma$ parameter was also adjusted, which defines the influence of each data point in the decision function. For DT and RF, the $max_depth$ parameter was tuned to limit the depth of the trees, controlling model complexity and preventing overfitting. The tested values were selected to cover a wide range of scenarios, ensuring that both simple and complex models were evaluated, and enabling the selection of the most suitable hyperparameters according to the specific characteristics of each dataset.

\begin{table*}[h]
  \centering
  \caption{Optimized hyperparameters and values tested for each algorithm.}
  \vspace{0.5cm}
  \begin{tabular}{|l|l|l|}
    \hline
    \textbf{Algorithm}                    & \textbf{Hyperparameter} & \textbf{Tested Values}                        \\
    \hline
    Logistic Regression (LR)              & $C$                     & [0.003, 0.03, 0.3, 3.0, 30.0]                 \\
    \hline
    Support Vector Machines (SVM)         & $C$                     & [0.3, 3.0, 30.0, 300.0]                       \\
    \cline{2-3}
                                          & $\gamma$                & [0.00003, 0.0003, 0.003, 0.03, 0.3]            \\
    \hline
    Random Forest (RF)                    & max\_depth              & [1, 5, 10, 15, 50]                            \\
    \hline
    Decision Tree (DT)                    & max\_depth              & [1, 5, 10, 15, 50]                            \\
    \hline
  \end{tabular}
  \label{tbl:hiper}
\end{table*}

These hyperparameters were then kept fixed throughout the entire experimental process to ensure that any variation in performance could be attributed solely to the cross-validation technique employed, rather than to subsequent hyperparameter adjustments. By its nature, Random Forest (RF) presents intrinsic fluctuations due to randomness in sample selection and tree construction, which may slightly impact the results of this algorithm. This procedure was essential to minimize external influences of hyperparameters on the measurements, enabling a more accurate assessment of the bias and variance characteristics of each cross-validation technique.

\subsection{Clustering Algorithms and Hyperparameter Definition}

The clustering algorithms used in the cross-validation procedures during the experiments require hyperparameters to be provided as input. While K-Means and Agglomerative Clustering require the number of clusters, DBSCAN depends on the parameters $\varepsilon$ and \(min\_samples\).

Thus, prior to the experiments, the clustering algorithm hyperparameters were estimated based on calculations. Following the same strategy proposed by Diamantidis et al. \cite{Diamantidis}, the number of clusters for the K-Means and Agglomerative Clustering algorithms was determined through repeated applications of hierarchical clustering on small samples of the datasets, using a similarity threshold between merged clusters to define the appropriate number of clusters. The resulting number of clusters for each dataset is presented in Table \ref{tbl:datasets}.

For the DBSCAN algorithm, the procedure suggested by Sander et al. \cite{sander1998gdbscan} was followed. The \(min_samples\) parameter, which defines the minimum number of neighboring points required to form a cluster, was set as \(2*\text{number of columns}\) in the dataset. The $\varepsilon$ parameter, which defines the radius within which data points are considered neighbors, was calculated using a \(k\)-distance graph, where the distance to the \(k\)-th nearest neighbor, with \textit{k} = \textit{min\_samples} - 1, is plotted in descending order. The appropriate value for $\varepsilon$ corresponds to the point at which a significant change in the slope of the graph occurs.

In the SCBCV Mini and KCBCV Mini algorithms, which employ Mini-Batch K-Means, the hyperparameter \(batch_size\) was set to the default value of \(1024\). The batch size determines the number of samples processed in each iteration of the algorithm, directly affecting both computational efficiency and clustering accuracy. A larger batch size allows more data to be processed at once, which can accelerate the algorithm’s convergence, but also requires more memory and may result in reduced accuracy for smaller datasets.

\subsection{Class-Stratified K-Means Cross-Validation}

The cross-validation technique proposed in this study, K-Means Cross-Validation with Class Stratification (SCBCV), was implemented based on two foundational methods: K-Means-based cross-validation and stratified cross-validation by class. Its pseudocode is presented in Algorithm \ref{alg:scbcv}.

In its input parameters, SCBCV uses the dataset $X$, containing the features to be analyzed, and $y$, which represents the corresponding class labels. The parameter $k_splits$ defines the number of folds for cross-validation, while $k_clusters$ specifies the number of clusters to be formed by the K-Means algorithm. The $rng$ parameter ensures experiment reproducibility through a random number generator, and $minibatch_kmeans$ indicates whether the optimized Mini-Batch K-Means version will be used to efficiently handle large volumes of data.

The SCBCV algorithm begins by organizing the input data $X$ and the corresponding class labels $y$, sorting and splitting them by class (line 2). Then, from lines 3 to 7, the algorithm checks whether to use Mini-Batch K-Means or the traditional K-Means algorithm, depending on the value of the $minibatch_kmeans$ parameter, to cluster the data into $k_clusters$.

After this setup, the algorithm proceeds to iterate over each class (line 9), applying K-Means to transform the class instances into distance vectors relative to the centroids (line 10), and retrieving the cluster index for each data point (line 11). From lines 13 to 16, all elements of the class are iterated over and assigned to their respective clusters. For each data point, an element is created associating its index with its distance to the centroid of its assigned cluster (line 14). This element is then appended to the corresponding cluster’s list (line 15) based on the previously determined cluster index. Each resulting cluster is then sorted by the distance of its elements to the centroid (line 18), and the sorted elements are added to a global index list called $index_list$ (line 19). Finally, in line 22, the elements in $index_list$ are distributed in a round-robin fashion across the $k_splits$ to form the different cross-validation folds, and these folds are returned as the algorithm’s output in line 23. This procedure ensures that each fold maintains a balanced data structure, with data points from the same class and belonging to the same cluster being distributed according to their distance to the centroid, that is, according to their similarity.

\floatname{algorithm}{Algorithm}
\begin{algorithm}
\caption{SCBCV}
\begin{algorithmic}[1]
    \Procedure{SCBCV}{$X$, $y$, $k\_splits$, $k\_clusters$, $rng$, $minibatch\_kmeans$}

        \State $X, y \gets$ orders  e divide $X, y$ por classe

        \If{$minibatch\_kmeans$}
            \State $kmeans \gets$ \textbf{MiniBatchKMeans($k\_clusters$, $rng$)}
        \Else
            \State $kmeans \gets$ \textbf{KMeans($k\_clusters$, $rng$)}
        \EndIf

        \State $index\_list \gets []$

        \For{\textbf{classe} em $X$}
            \State $class\_dist \gets$ executa $kmeans$ e retorna matriz de distâncias de cada ponto a cada centróide
            \State $clusters\_index \gets$ obtém índices dos clusters de cada elemento em $class\_dist$
            \State $clusters \gets$ listas vazias para cada um dos $k\_clusters$
            
            \For{\textbf{index} em $class\_dist$}
                \State $elemento \gets$ (index, distância ao centróide)
                \State $clusters[clusters\_index[index]] \gets elemento$
            \EndFor

            \For{\textbf{cluster} em $clusters$}
                \State ordena cluster por distância ao centróide
                \State $index\_list \gets index\_list + cluster$ ordenado
            \EndFor
        \EndFor
        \State $folds \gets$ distribui $index\_list$ entre $k\_splits$ $folds$ de forma circular

        \State \textbf{return} $folds$
    \EndProcedure
\end{algorithmic}
\label{alg:scbcv}
\end{algorithm}

\subsection{Evaluation Metrics}

To evaluate the performance of the cross-validation techniques in the experiment, it is essential to use metrics that provide a comprehensive understanding of the quality of the produced estimates. The metrics chosen in this study were bias and variance, following prior works of similar nature, such as those by Kohavi \cite{kohavi1995study} and Cawley et al. \cite{cawley2010bias}.

Bias measures the difference between the expected estimate of the model’s performance and its true performance, allowing for the evaluation of how much the cross-validation technique may underestimate (when negative) or overestimate (when positive) the model’s predictive ability. Variance, on the other hand, quantifies the sensitivity of the validation technique to variations in the training data, indicating the stability of the obtained estimates. By combining these two metrics, a balanced view of the performance of the tested cross-validation techniques can be achieved.

For balanced datasets, accuracy was used as the performance metric, which evaluates the number of correct predictions over the total number of predictions made, as shown in Equation \ref{eq:2}. This metric is reliable when classes contain similar amounts of data. However, in imbalanced datasets, accuracy often fails to reflect the model's ability to identify instances of the minority class. Therefore, for imbalanced datasets, the F1-score was used as the performance metric, as shown in Equation \ref{eq:3}. The F1-score is preferable in such cases because it is the harmonic mean of precision and recall (or sensitivity).

\begin{equation}
\label{eq:2}
\text{Accuracy} = \frac{TP + TN}{TP + TN + FP + FN}
\end{equation}

\begin{equation}
\label{eq:3}
\text{F1-score} = 2 \times \frac{\text{Precision} \times \text{Recall}}{\text{Precision} + \text{Recall}}
\end{equation}

In Equation \ref{eq:2}, TP (True Positives) and TN (True Negatives) refer to the number of positive instances correctly classified as positive by the model, and the number of negative instances correctly classified as negative, respectively. FP (False Positives) and FN (False Negatives), in turn, represent the number of negative instances incorrectly classified as positive, and positive instances incorrectly classified as negative, respectively.

In Equation \ref{eq:3}, Precision is the proportion of correct predictions for the positive class relative to all predictions made for that class, while Recall is the proportion of actual positive instances that were correctly identified by the model.

Since this study uses real-world datasets, obtaining the true test performance is unfeasible. However, it is possible to estimate it using a high number of stratified holdout repetitions, as done by Budka et al. \cite{budka2013density} A total of 100 repetitions of this strategy were conducted, with 90\% of each dataset used for training, and the results were averaged. A large training set was chosen to reduce bias, which could occur if a smaller training set were used, while the variance is expected to be mitigated by the high number of holdout repetitions.

The expected estimate of each cross-validation technique was calculated for each dataset by sampling 90\% of the dataset 20 times and applying the cross-validation technique to obtain estimates of the true performance. The average of these 20 estimates was used as the expected performance estimate to be approximated by the cross-validation method. Thus, considering \( CV_i \) as the performance estimate obtained by applying K-fold cross-validation on a given dataset using one of the cross-validation techniques, the expected estimate of cross-validation is given by Equation \ref{eq:4}.

\begin{equation}
\label{eq:4}
CV = \frac{1}{20} \sum_{i=1}^{20} CV_i
\end{equation}

Bias is therefore computed using  \( b_{CV} = CV - \hat{P} \), where \( \hat{P} \) is the estimate of the true performance obtained via the 100-fold repeated stratified hold-out procedure described above.

Variance is calculated according to Equation \ref{eq:5}. For readability, this study reports the standard deviation (\( s \)) instead of the variance, as the two are directly related. Bias and variance were evaluated for the various cross-validation strategies across 20 different datasets and four learning algorithms. For each K-fold cross-validation strategy, experiments were conducted with both 2 and 10 folds.

\begin{equation}
\label{eq:5}
s^2_{CV} = \frac{1}{20 - 1} \sum_{i=1}^{20} (CV_i - CV)^2
\end{equation}

Finally, the computational cost of each cross-validation technique was measured with \textit{perf\_counter()} from the Python library \textit{time}. The clock was recorded immediately before and after each cross-validation run, providing an accurate measure of execution time, and hence the computational cost, associated with every technique tested.

\subsection{Statistical Analysis}

In this study, the Friedman test was used to assess the statistical significance of performance differences among the various cross-validation techniques applied in the experiments. The Friedman test is a non-parametric test used when comparing three or more treatments or algorithms under similar conditions, where the data consist of repeated measures across blocks—as is the case in cross-validation experiments \cite{friedman1940comparison}.

For each set of experiments, the Friedman test was applied separately to the bias and standard deviation metrics. The test evaluates the null hypothesis that all cross-validation techniques have equivalent performance. If the resulting  \textit{p}-value is below a predefined significance level (typically 0.05), the null hypothesis is rejected, indicating that at least one technique performs significantly differently from the others.

\subsection{Practical Implementation}

The implementation of the algorithms and experiments was developed in Python using the Scikit-Learn library, which provides a broad range of tools for machine learning and data analysis. The environment setup also incorporated auxiliary libraries: NumPy for array manipulation, Pandas for data handling, SciPy for conducting the Friedman tests, and Matplotlib for visualising the plots produced by the Kneed library for determining the $\varepsilon$ parameters.

For the cross-validation experiments, the clustering algorithms K-Means, Mini-Batch K-Means, DBSCAN, and Agglomerative Clustering—all available in Scikit-Learn—were employed. Hyperparameter selection for these algorithms followed the methods described earlier, with values adjusted to the specific characteristics of each dataset. The code was structured in a modular fashion, allowing flexible replication and modification of the experiments while ensuring reproducibility of the results. All experiments were executed on a multi-core machine, leveraging parallelisation to optimise runtime, particularly in the phases involving repeated execution of the clustering algorithms and cross-validation.

The experimental framework was based on the work of Fontanari et al. \cite{Fontanari}, with modifications to the parallelisation structures, cluster parameter settings, dataset selection, and experimental design. The complete code for this study is available at \url{https://github.com/amspezia/K-Fold-Partitioning-Methods}.

\section{Results and Discussion}
\label{sec:results}

The experiments involved applying the cross-validation techniques to each of the 20 datasets and 4 learning algorithms, resulting in a total of 80 bias and variance samples per technique. The value of \textit{k} in the K-fold method was set to 2 and 10, representing two distinct scenarios: one with a smaller number of folds, which is more susceptible to variability, and another with a larger number of folds, which balances bias and variance more effectively. Below, each of the experimental sets defined in Section~\ref{sec:methodology} is described along with the results obtained.

The experiments were conducted on a notebook equipped with a 13th Gen Intel(R) Core(TM) i7-1355U 1.70 GHz processor, with 10 cores and 16 GB of RAM. To optimize performance, the execution of each cross-validation routine was parallelized across all processor cores, ensuring efficient use of available computational resources. The results were then aggregated for final analysis.

\subsection{Set 1}

The primary objective of this experimental set was to determine how the number of clusters, used as a parameter in the SCBCV cross-validation technique, influences its results. In SCBCV the cluster count is applied within each class—rather than to the dataset as a whole—because the method combines class stratification with the K-Means clustering algorithm.

As discussed in the previous section, the cluster numbers for K-Means and Agglomerative Clustering were estimated for each dataset following Diamantidis et al. \cite{Diamantidis}. That same dataset-specific value was then used in SCBCV and compared with fixed settings of 2, 3, 4, and 5 clusters. These runs are denoted SCBCV (variable value), and SCBCV2, SCBCV3, SCBCV4, and SCBCV5 (fixed values).

The \textit{p}-values from the Friedman tests for Set 1 (Table \ref{tbl:1frid}) show that, across the datasets and metrics evaluated, there were no statistically significant differences in SCBCV performance when different cluster counts were used. For both bias and standard deviation, all \textit{p}-values exceeded the conventional 0.05 threshold.

\begin{table}[h]
  \centering
  \vspace{0.5cm}
  \caption{P-values of bias and standard deviation from Friedman tests conducted on Experiment Set 1.}
  \label{tbl:1frid}
  \begin{tabular}{|c|c|c|c|c|}
    \hline
    \textbf{Metric} & \textbf{Balancing} & \textbf{Folds} & \multicolumn{2}{c|}{\textbf{p-value}} \\
    \cline{4-5}
                    &                    &                 & \textbf{bias}  & \textbf{std.\ dev.} \\
    \hline
    \multirow{2}{*}{Accuracy}   & \multirow{2}{*}{Balanced}   & 2   & 0.29901 & 0.89176 \\
    \cline{3-5}
                                &                             & 10  & 0.20231 & 0.44294 \\
    \hline
    \multirow{2}{*}{F1-score}   & \multirow{2}{*}{Imbalanced} & 2   & 0.82845 & 0.92485 \\
    \cline{3-5}
                                &                             & 10  & 0.86324 & 0.46102 \\
    \hline
  \end{tabular}
\end{table}

\begin{figure*}[tph]
  \centering
    \includegraphics[width=0.70\columnwidth,valign=c]%, 
        {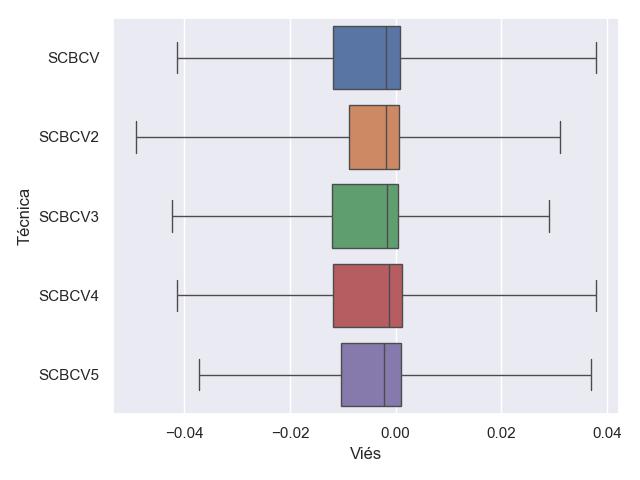}%
    \hspace{0.1\textwidth}%
    \includegraphics[width=0.70\columnwidth,valign=c]%
        {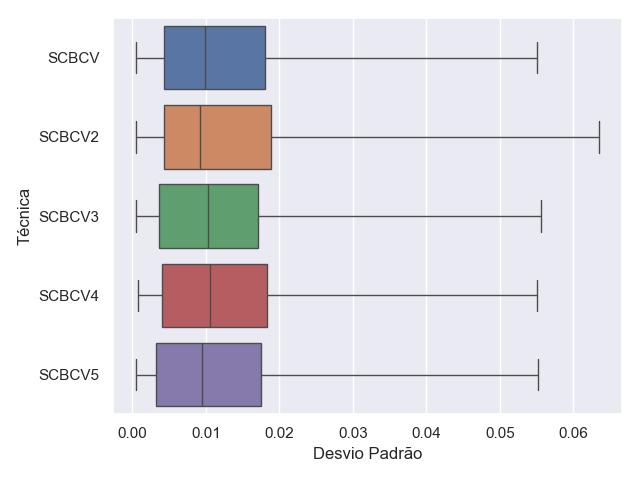}%
    \caption{Mean bias (a) and standard deviation (b) of the SCBCV 10-fold procedure on Experiment Set 1, for balanced datasets.}
    \label{fig:1balanced10}
\end{figure*}

Figure \ref{fig:1balanced10} plots mean bias and standard deviation for each technique with 10-fold cross-validation on balanced datasets. SCBCV2 exhibits the bias closest to zero but also the largest standard deviation. Conversely, SCBCV3 shows a slightly higher bias yet a lower standard deviation. Overall, the techniques perform very similarly when both bias and variability are considered.

\begin{figure*}[tph]
  \centering
    \includegraphics[width=0.70\columnwidth,valign=c]%
      {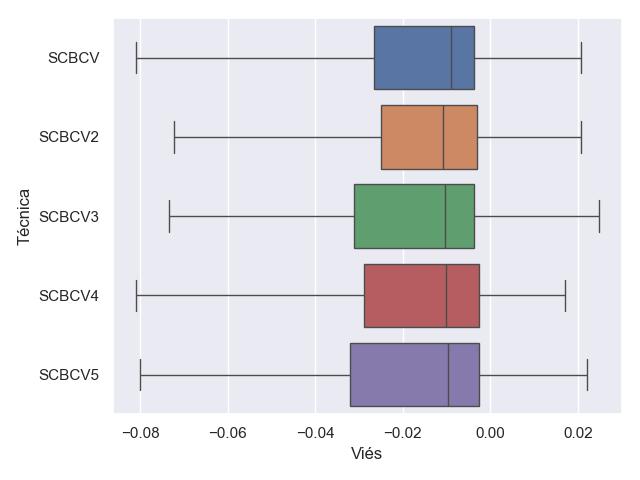}%
    \hspace{0.1\textwidth}%
    \includegraphics[width=0.70\columnwidth,valign=c]%
      {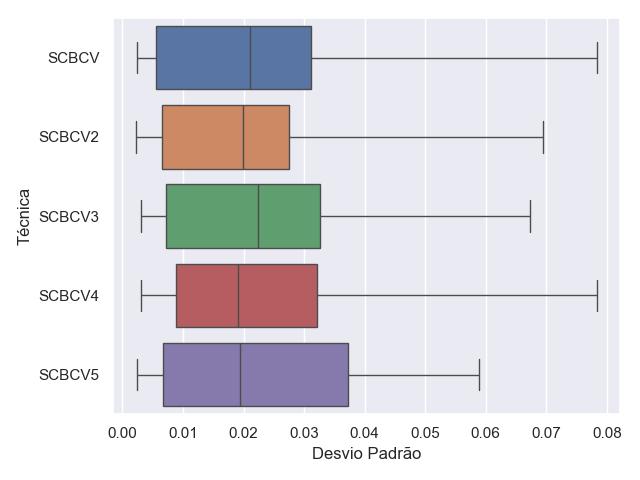}%
  \caption{Mean bias (a) and standard deviation (b) of the SCBCV 10-fold procedure on Experiment Set 1, for imbalanced datasets.}
  \label{fig:1imbalanced10}
\end{figure*}

\begin{figure*}[tph]
    \centering
        \includegraphics[width=0.70\columnwidth,valign=c]{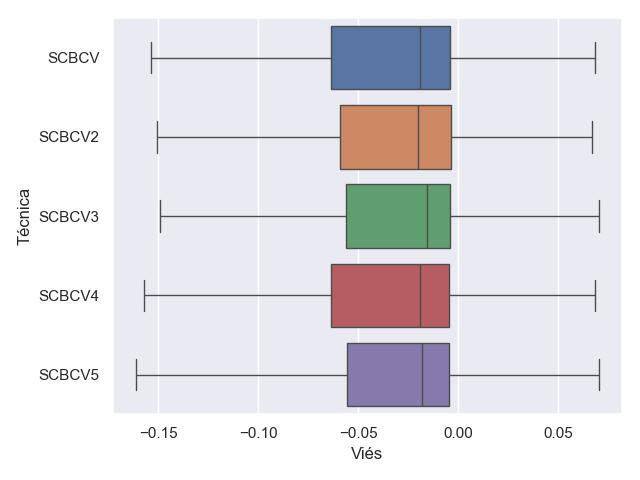}%
    \hspace{0.1\textwidth}%
        \includegraphics[width=0.70\columnwidth,valign=c]%
        {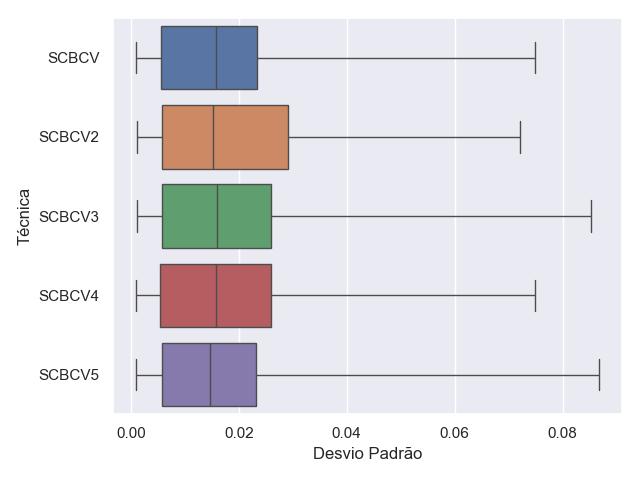}%
    \caption{Mean bias (a) and standard deviation (b) of the SCBCV 2-fold procedure on Experiment Set 1, for balanced datasets.}
    \label{fig:1balanced2}
\end{figure*}

\begin{figure*}[tph]
    \centering
        \includegraphics[width=0.70\columnwidth,valign=c]{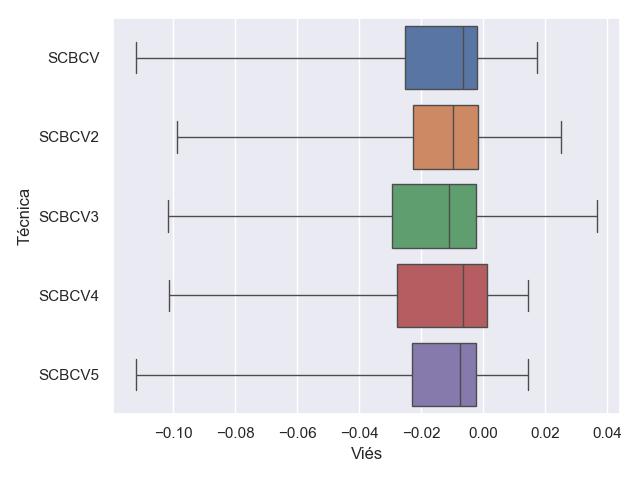}
    \hspace{0.1\textwidth}%
        \includegraphics[width=0.70\columnwidth,valign=c]
        {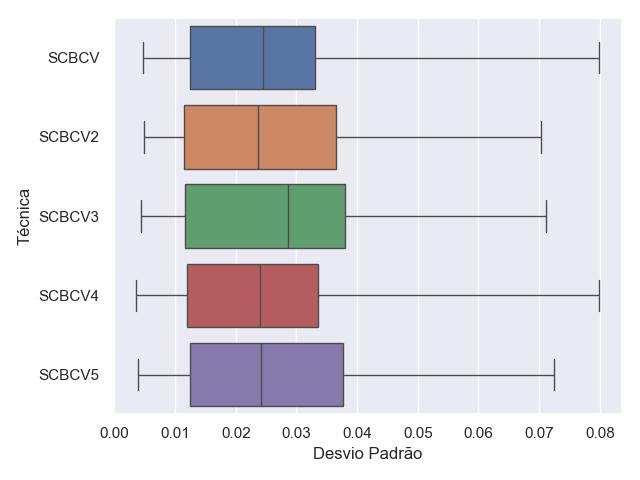}
    \caption{Mean bias (a) and standard deviation (b) of the SCBCV 2-fold procedure on Experiment Set 1, for imbalanced datasets.}
    \label{fig:1imbalanced2}
\end{figure*}

In Figure \ref{fig:1imbalanced10}, the same 10-fold experiment is applied to imbalanced datasets. SCBCV stands out with the median bias closest to zero, followed by SCBCV2, which has the lowest mean standard deviation. The other techniques show higher values for these metrics. However, in the experiments using 2 folds, as shown in Figures \ref{fig:1balanced2} and \ref{fig:1imbalanced2}, SCBCV2 does not maintain the same performance, and SCBCV proves to be the most stable technique compared to the others.

\begin{figure}[tph]
    \begin{center}
        \includegraphics[width=0.90\columnwidth]{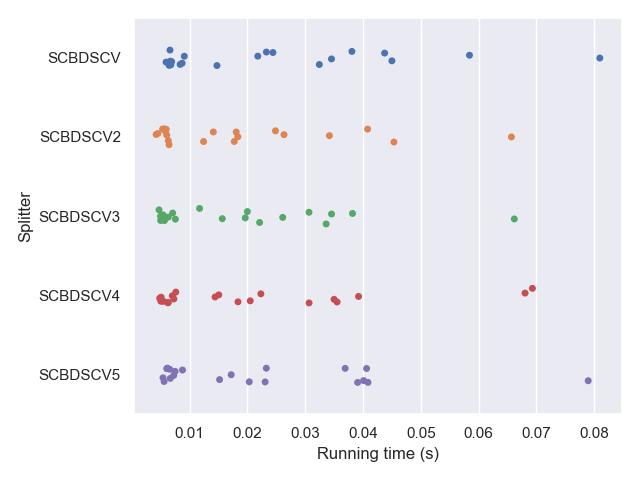}
    \end{center}
    
    \caption{Overall execution time of the techniques from Experiment Set 1 on 20 selected datasets using 10 folds.}
    \label{fig:1runningtimes}
\end{figure}

In terms of computational cost, all runs of the technique exhibited very similar execution times. As the number of clusters increases, the average execution time also grows, as can be seen in Figure \ref{fig:1runningtimes}. The SCBCV variant with a variable number of clusters showed an average time comparable to the others, slightly slower, but not to the extent of being a significant drawback. In summary, the impact of the number of clusters on SCBCV’s execution time is noticeable yet very small, which reinforces the feasibility of using this technique even in scenarios where computational cost is a concern.

In summary, the impact of the number of clusters on SCBCV’s runtime is noticeable but minimal, which reinforces the practical feasibility of using this technique even in scenarios where computational cost is a concern.

Overall, and in line with the Friedman test results, the plots presented in the figures suggest that the SCBCV technique is relatively insensitive to the number of clusters used—whether determined by the method of Diamantidis et al. \cite{Diamantidis} or fixed at 2, 3, 4, or 5 clusters. While minor variations exist, no single configuration consistently outperforms the others by a significant margin.

This insensitivity indicates that the algorithm can be used flexibly, without the strict need to optimize the number of clusters for each dataset. The robustness observed concerning the number of clusters reflects the stability of the algorithm.

Given that the SCBCV variant using a predefined, dataset-specific number of clusters demonstrated the most consistent performance across multiple settings, it was selected for use in the subsequent experiments.

\subsection{Set 2}

This set of experiments aims to compare the SCBCV technique using traditional K-Means, the SCBCV Mini technique using Mini-Batch K-Means, and the well-established Stratified Cross-Validation, referred to here as SCV. The experimental setup mirrored that of the previous set, applying each cross-validation method to the 20 datasets listed in Table \ref{tbl:datasets}, across the four learning algorithms presented earlier.

The Friedman analysis conducted on the data from Set 2 (Table \ref{tbl:2frid}) suggests that, for the accuracy metric in balanced datasets, the bias and standard deviation values in both 2-fold and 10-fold settings do not show statistically significant differences, as all p-values exceed 0.05. This indicates a relative stability in the performance of the cross-validation techniques across these datasets, regardless of the number of folds used.

\vspace{0.5cm}      
\begin{table}[tph]
\centering
\caption{P-values of bias and standard deviation from Friedman tests conducted on Experiment Set 2.}
\vspace{0.5cm}
\begin{tabular}{|c|c|c|c|c|}
\hline
\textbf{Metric} & \textbf{Dataset Type} & \textbf{Folds} & \multicolumn{2}{c|}{\textbf{p-value}} \\
\cline{4-5}
                &                       &                 & \textbf{bias}  & \textbf{std.\ dev.} \\
\hline
\multirow{2}{*}{Accuracy}   & \multirow{2}{*}{Balanced}    & 2   & 0.09778  & 0.07261 \\
\cline{3-5}
                             &                              & 10  & 0.90484  & 0.74082 \\
\hline
\multirow{2}{*}{F1-score}   & \multirow{2}{*}{Imbalanced}  & 2   & 0.20190  & 0.01357 \\
\cline{3-5}
                             &                              & 10  & <0.00001 & 0.01488 \\
\hline
\end{tabular}
\label{tbl:2frid}
\end{table}

For the F1 metric in imbalanced datasets, the p-values indicate a more pronounced difference between the executions with 2 and 10 folds. With \textit{p}-values near zero for bias with 10 folds and below 0.02 for standard deviation in both 2 and 10-fold scenarios, the Friedman test reveals statistically significant differences in these datasets. These results suggest that both bias and standard deviation estimates are influenced by the choice of cross-validation technique.

\begin{figure*}[tph]
    \centering
        \includegraphics[width=0.70\columnwidth,valign=c]%
        {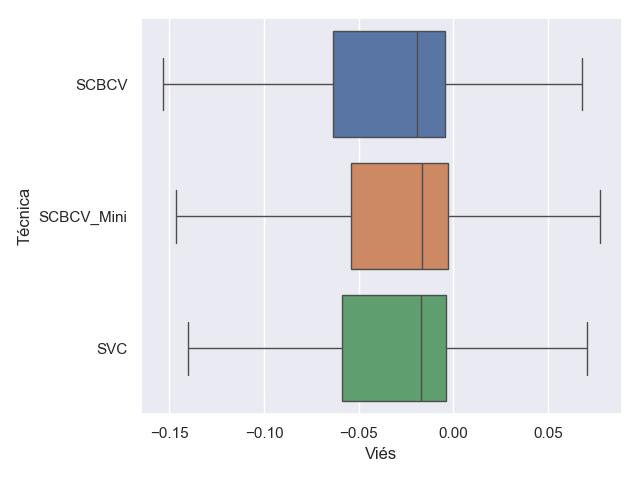}%
    \hspace{0.1\textwidth}%
        \includegraphics[width=0.70\columnwidth,valign=c]%
        {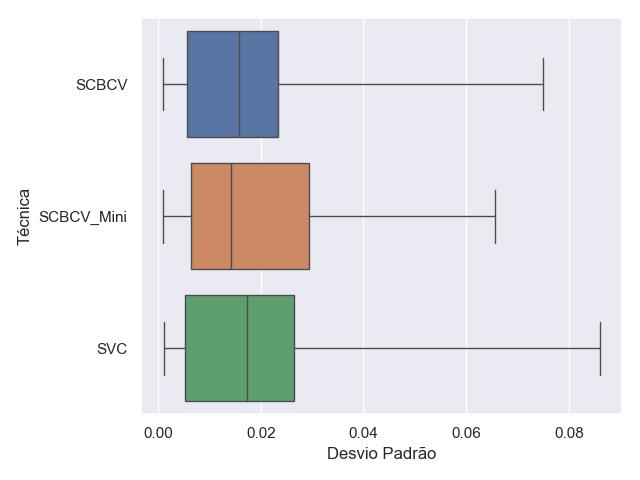}%
    \caption{Mean bias (a) and standard deviation (b) of the SCBCV, SCBCV Mini and SCV 2-fold techniques in Experiment Set 2, for balanced datasets.}
    \label{fig:2balanced2}
\end{figure*}

In Figure \ref{fig:2balanced2}, the similarity in performance across the techniques is evident in terms of both bias and standard deviation when using 2-fold cross-validation on balanced datasets. In this setting, SCBCV Mini stands out with the median bias closest to zero and the lowest median standard deviation. On the other hand, SCV, while showing the smallest spread in bias values, exhibits the greatest spread in standard deviation, indicating more variability in its error estimates.

\begin{figure*}[tph]
    \centering
        \includegraphics[width=0.70\columnwidth,valign=c]%
        {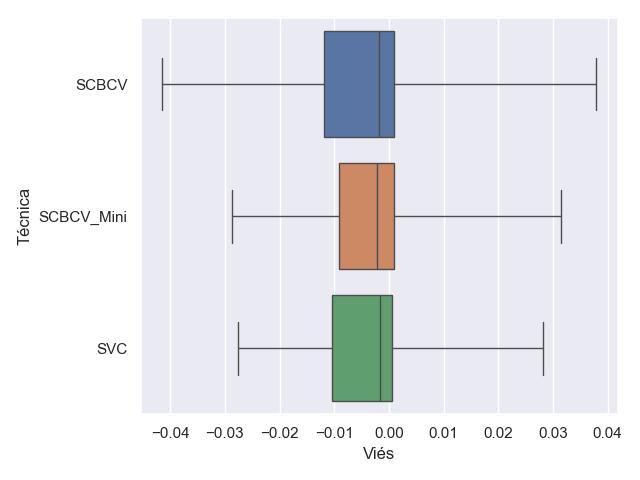}%
    \hspace{0.1\textwidth}%
        \includegraphics[width=0.70\columnwidth,valign=c]%
        {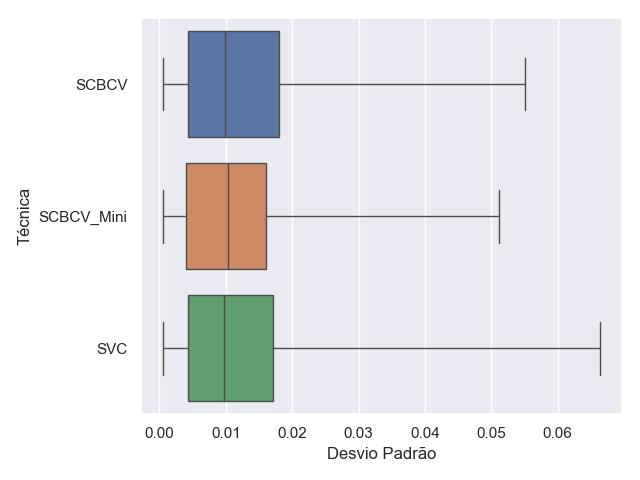}%
    \caption{Mean bias (a) and standard deviation (b) of the SCBCV, SCBCV Mini, and SCV 10-fold techniques in Experiment Set 2, for balanced datasets.}
    \label{fig:2balanced10}
\end{figure*}

Regarding balanced datasets with 10 folds (Figure \ref{fig:2balanced10}), SCV exhibits slightly lower median bias and standard deviation than SCBCV Mini, although the bulk of its metric values lie farther from zero. SCBCV shows the greatest dispersion in bias values, making it less competitive than its Mini-Batch counterpart.

\begin{figure*}[tph]
    \centering
        \includegraphics[width=0.70\columnwidth,valign=c]%
        {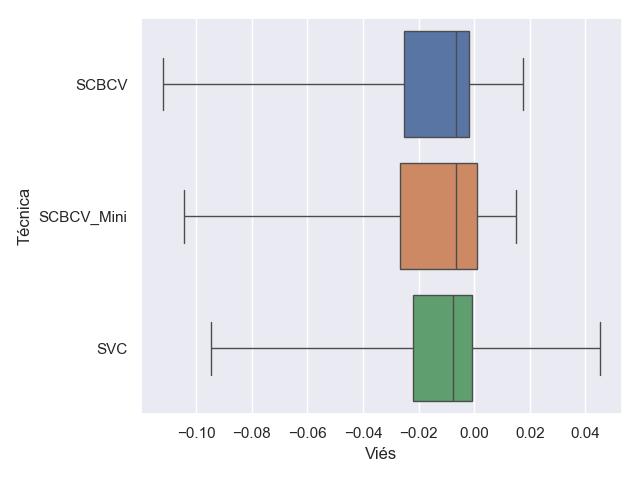}%
    \hspace{0.1\textwidth}%
        \includegraphics[width=0.70\columnwidth,valign=c]%
        {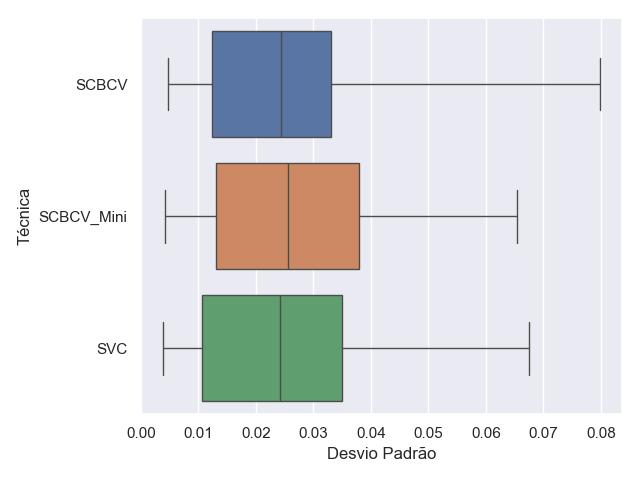}%
    \caption{Mean bias (a) and standard deviation (b) of the SCBCV, SCBCV Mini, and SCV 2-fold techniques in Experiment Set 2, for imbalanced datasets.}
    \label{fig:2imbalanced2}
\end{figure*}

\begin{figure*}[tph]
    \centering
        \includegraphics[width=0.70\columnwidth,valign=c]%
        {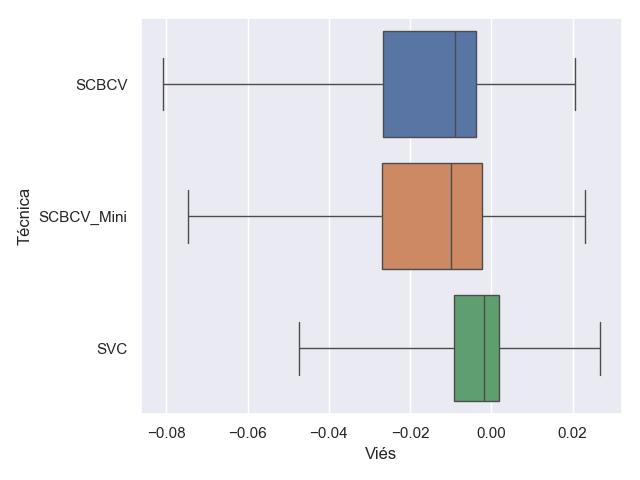}%
    \hspace{0.1\textwidth}%
        \includegraphics[width=0.70\columnwidth,valign=c]%
        {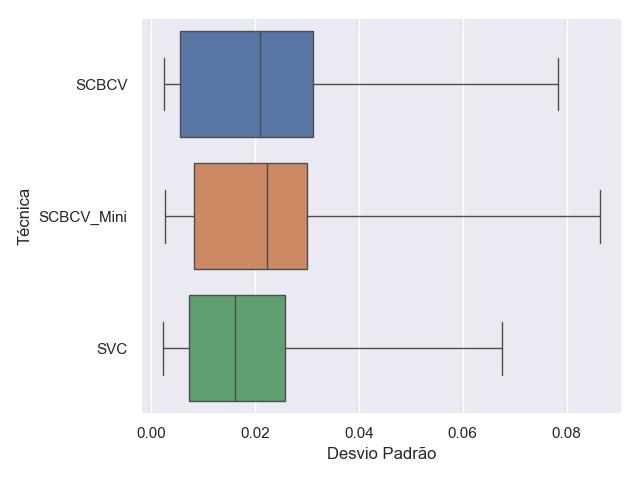}%
    \caption{Mean bias (a) and standard deviation (b) of the SCBCV, SCBCV Mini, and SCV 10-fold techniques in Experiment Set 2, for imbalanced datasets.}
    \label{fig:2imbalanced10}
\end{figure*}

For imbalanced datasets, the differences among techniques become more pronounced. In Figure \ref{fig:2imbalanced2}, SCV stands out modestly by producing standard-deviation values closer to zero when only 2 folds are used, yet it also yields more positive bias values, indicating a tendency to overestimate model performance compared with cluster-based methods. SCV’s advantage grows markedly by 10-fold (Figure \ref{fig:2imbalanced10}): the medians of both bias and standard deviation for SCV are visibly closer to zero than those of the SCBCV variants, and its overall range of variability is considerably narrower.

Table \ref{tbl:2wins} highlights how often each cross-validation technique achieved the best performance for each dataset type and metric. The results confirm that SCV outperforms the other techniques in imbalanced datasets, aligning with the preliminary findings from the Friedman tests and the trends observed in the figures. In balanced datasets, although the Friedman tests indicate that differences among the techniques are not statistically significant, the plots show a slight advantage for SCBCV Mini, which is further supported by the data in Table \ref{tbl:2wins}.

\begin{table*}[tph]
  \centering
  \caption{Number of times each validation technique tested in Experiment Set 2 achieved the best result in terms of bias and standard deviation. The winning technique in each row is highlighted.}
  \vspace{0.5cm}
    \begin{tabular}{|c|c|c|c|c|c|c|}
      \hline
      \textbf{Metric} & \textbf{Dataset Type} & \textbf{Folds} & \textbf{Measure}       & \textbf{SCBCV} & \textbf{SCBCV Mini} & \textbf{SCV} \\ 
      \hline
      \multirow{4}{*}{\textbf{Accuracy}} 
        & \multirow{4}{*}{Balanced} 
        & \multirow{2}{*}{2} 
          & Bias             & 13 & \textbf{16} & 11 \\ \cline{4-7}
        & 
        & 
          & Std.\ Dev.       & 16 & \textbf{19} & 5  \\ \cline{3-7}
        & 
        & \multirow{2}{*}{10}
          & Bias             & 9  & \textbf{16} & 15 \\ \cline{4-7}
        & 
        & 
          & Std.\ Dev.       & \textbf{15} & 14 & 11 \\ 
      \hline
      \multirow{4}{*}{\textbf{F1-score}}
        & \multirow{4}{*}{Imbalanced}
        & \multirow{2}{*}{2}
          & Bias             & \textbf{18} & 11 & 11 \\ \cline{4-7}
        & 
        & 
          & Std.\ Dev.       & 14 & 6  & \textbf{20} \\ \cline{3-7}
        & 
        & \multirow{2}{*}{10}
          & Bias             & 5  & 5  & \textbf{30} \\ \cline{4-7}
        & 
        & 
          & Std.\ Dev.       & 15 & 6  & \textbf{19} \\ 
      \hline
    \end{tabular}
  \label{tbl:2wins}
\end{table*}

\begin{figure}[tph]
    \begin{center}
        \includegraphics[width=0.90\columnwidth]{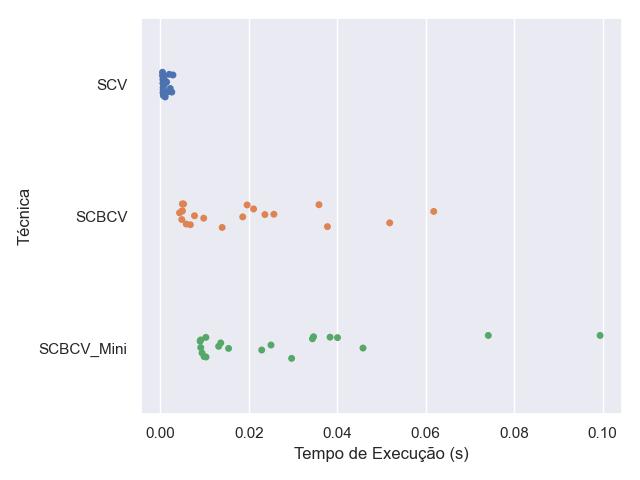}
    \end{center}
    \caption{Overall execution time of the techniques in Experiment Set 2 on the 20 datasets using 10 folds.}
    \label{fig:2runningtimes}
\end{figure}

Finally, Figure \ref{fig:2runningtimes} shows the execution time of each technique across the different datasets. SCV stands out as the fastest technique, confirming findings from previous studies that compared SCV to cluster-based cross-validation methods \cite{Fontanari}. However, contrary to expectations prior to the experiments, SCBCV using Mini-Batch K-Means was slower than SCBCV with traditional K-Means. This suggests that, although Mini-Batch K-Means is theoretically more efficient for large-scale data processing, its performance advantage may not hold in all scenarios.

The results obtained in this set of experiments suggest that while SCV is the most robust choice for imbalanced scenarios, SCBCV Mini can be considered an effective alternative for balanced datasets. Therefore, the choice of cross-validation technique should be guided by the dataset characteristics and the performance metric most relevant to the model under evaluation.

\subsection{Set 3}

This set of experiments aims to compare various cluster-based cross-validation techniques, specifically SCBCV, SCBCV Mini, KCBCV, KCBCV Mini, ACBCV, and DBSCANBCV. The inclusion of these techniques allows for the evaluation not only of the effectiveness of traditional K-Means and Mini-Batch K-Means—as explored in the previous sets—but also an extension of the analysis to other clustering algorithms, such as Agglomerative Clustering (ACBCV) and DBSCAN (DBSCANBCV). This comparison seeks to assess the impact of the clustering algorithm choice on performance estimation and to determine whether any method offers a clearly superior balance between bias, variance, and execution time across scenarios with varying data distributions and complexities.

Table \ref{tbl:3frid} presents the \textit{p}-values from the Friedman tests for the bias and standard deviation data obtained in Set 3. It shows that, for most metrics used, no statistically significant differences were observed between the techniques. However, in imbalanced datasets with 10 folds, the \textit{p}-value of 0.00422 for bias indicates that the choice of technique does tend to influence the results for this metric in that specific context.

\vspace{0.5cm}
\begin{table}[hpt]
  \centering
  \caption{P-values of bias and standard deviation from Friedman tests conducted on Experiment Set 3.}
  \vspace{0.5cm}
  \begin{tabular}{|c|c|c|c|c|}
    \hline
    \textbf{Metric}    & \textbf{Dataset Type} & \textbf{Folds} & \multicolumn{2}{c|}{\textbf{p-value}} \\
    \cline{4-5}
                       &                       &                & \textbf{bias} & \textbf{std.\ dev.} \\
    \hline
    \multirow{2}{*}{Accuracy} 
                       & \multirow{2}{*}{Balanced} & 2  & 0.24267  & 0.88889 \\
    \cline{3-5}
                       &                         & 10 & 0.31952  & 0.09580 \\
    \hline
    \multirow{2}{*}{F1-score} 
                       & \multirow{2}{*}{Imbalanced} & 2  & 0.15311  & 0.16582 \\
    \cline{3-5}
                       &                           & 10 & 0.00422  & 0.13075 \\
    \hline
  \end{tabular}
  \label{tbl:3frid}
\end{table}

\begin{figure*}[tph]
    \centering
        \includegraphics[width=0.70\columnwidth,valign=c]%
        {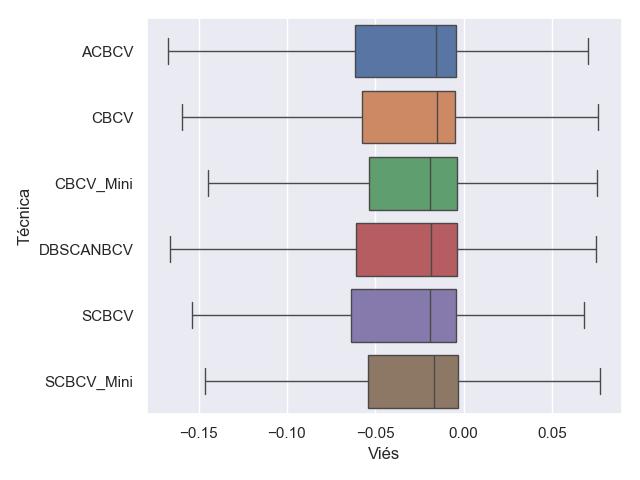}%
    \hspace{0.1\textwidth}%
        \includegraphics[width=0.70\columnwidth,valign=c]%
        {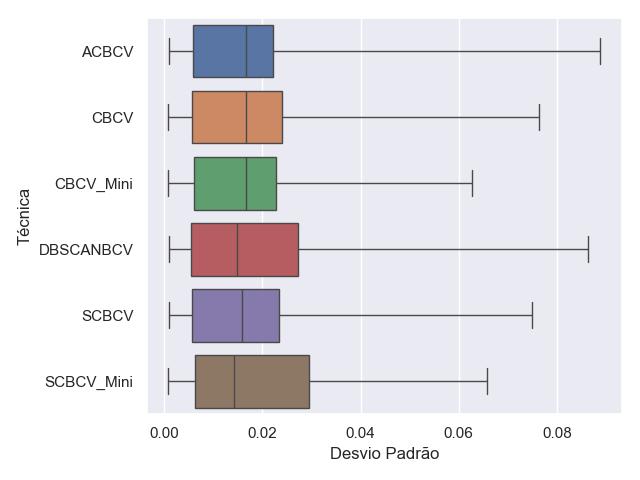}%
    \caption{Mean bias (a) and standard deviation (b) of the 2-fold cluster-based cross-validation techniques in Experiment Set 3, for balanced datasets.}
    \label{fig:3balanced2}
\end{figure*}

In Figure \ref{fig:3balanced2}, it can be observed that the various cluster-based cross-validation techniques exhibit very similar bias and standard deviation in balanced datasets using 2 folds. While KCBCV and ACBCV have bias values with medians closest to zero, DBSCANBCV and SCBCV Mini show the lowest median standard deviation values.

\begin{figure*}[tph]
    \centering
        \includegraphics[width=0.70\columnwidth,valign=c]%
        {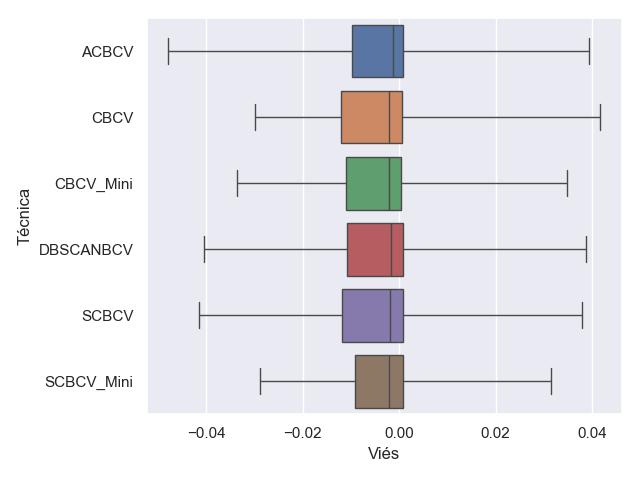}%
    \hspace{0.1\textwidth}%
        \includegraphics[width=0.70\columnwidth,valign=c]%
        {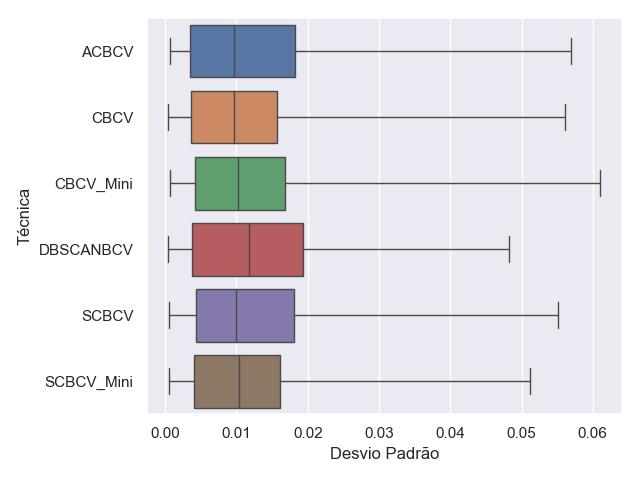}%
    \caption{Mean bias (a) and standard deviation (b) of the 10-fold cluster-based cross-validation techniques in Experiment Set 3, for balanced datasets.}
    \label{fig:3balanced10}
\end{figure*}

When using 10 folds in balanced datasets, as shown in Figure \ref{fig:3balanced10}, SCBCV Mini stands out in both metrics, as most of its values fall within ranges closer to zero compared to the other techniques. However, the differences among the methods remain minimal, reinforcing their comparable performance in such scenarios.

\begin{figure*}[tph]
    \centering
        \includegraphics[width=0.70\columnwidth,valign=c]%
        {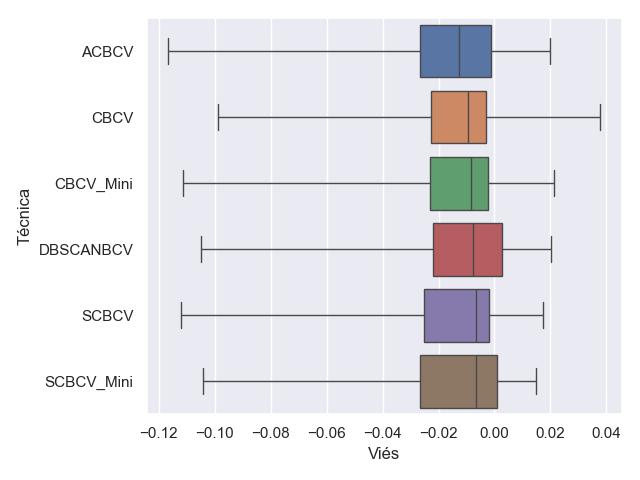}%
    \hspace{0.1\textwidth}%
        \includegraphics[width=0.70\columnwidth,valign=c]%
        {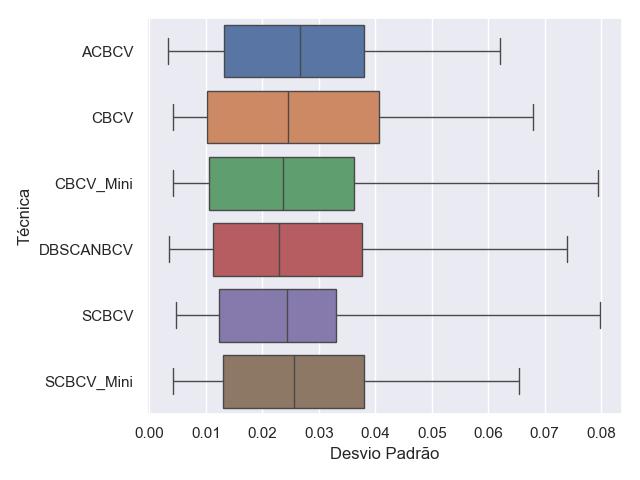}%
    \caption{Mean bias (a) and standard deviation (b) of the 2-fold cluster-based cross-validation techniques in Experiment Set 3, for imbalanced datasets.}
    \label{fig:3imbalanced2}
\end{figure*}

In imbalanced datasets with 2 folds (Figure \ref{fig:3imbalanced2}), the metric values are again very similar across methods. The standout here is DBSCANBCV, which shows bias values clustered closer to zero and the lowest median standard deviation. Finally, for imbalanced datasets with 10 folds, DBSCANBCV again appears to outperform the others, with bias and standard-deviation values nearer to zero. The SCBCV variants, however, display weaker performance under these conditions. Even so, the differences among techniques remain subtle.

\begin{figure*}[tph]
    \centering
        \includegraphics[width=0.70\columnwidth,valign=c]%
        {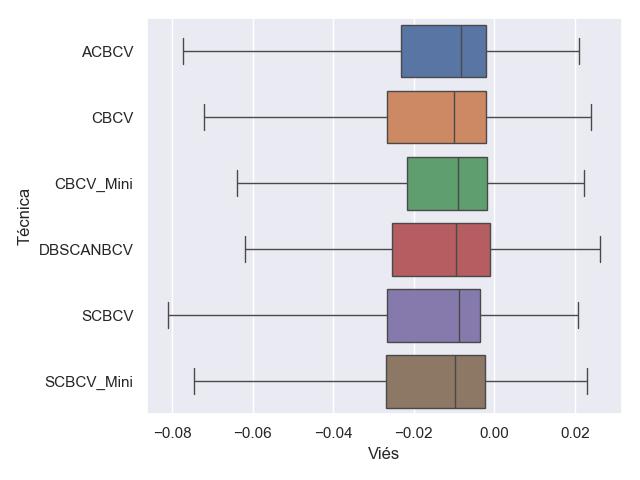}%
    \hspace{0.1\textwidth}%
        \includegraphics[width=0.70\columnwidth,valign=c]%
        {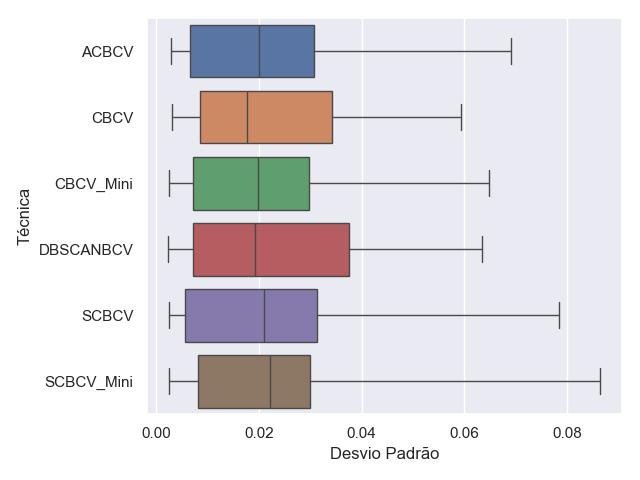}%
    \caption{Mean bias (a) and standard deviation (b) of the 10-fold cluster-based cross-validation techniques in Experiment Set 3, for imbalanced datasets.}
    \label{fig:3imbalanced10}
\end{figure*}

Table \ref{tbl:3wins}, which counts how many times each technique achieved the best result relative to the others, allows a deeper reading. SCBCV failed to lead in any of the tested scenarios, whereas ACBCV yielded the top results in balanced datasets. In imbalanced datasets, KCBCV was superior when only 2 folds were used. Although these counts are not deterministic, they help clarify how the algorithms perform under differing class-balance settings and fold counts.

\begin{table*}[h]
  \centering
  \caption{Number of times each validation technique tested in Experiment Set 3 achieved the best result in terms of bias and standard deviation. The winning technique in each row is highlighted. ‘Imbalanced’ has been abbreviated to ‘Imb.’, ‘Balanced’ to ‘Bal.’, ‘Standard deviation’ to ‘Std.\ Dev.’, and the term ‘BCV’ (Based Cross-Validation) has been removed for better data presentation.}
  \vspace{0.5cm}
  \begin{tabular}{|c|c|c|c|c|c|c|c|c|c|}
    \hline
           &         &       &         & \textbf{AC.} & \textbf{KC.} & \textbf{KC. Mini} & \textbf{DBSCAN.} & \textbf{SC.} & \textbf{SC. Mini} \\ 
    \hline
    \multirow{4}{*}{\textbf{Accuracy}} 
      & \multirow{2}{*}{Bal.} 
      & 2    
      & Bias       & 9  & 4  & 6  & 3  & 7  & \textbf{11} \\ 
    \cline{3-10}
      & 
      &       
      & Std.\ Dev. & \textbf{9}  & 3  & \textbf{9}  & \textbf{9}  & 5  & 5  \\ 
    \cline{3-10}
      & 
      & 10   
      & Bias       & \textbf{12} & 6  & 3  & 4  & 6  & 9  \\ 
    \cline{3-10}
      & 
      &       
      & Std.\ Dev. & 6  & \textbf{11} & 5  & 2  & 9  & 7  \\ 
    \hline
    \multirow{4}{*}{\textbf{F1-score}} 
      & \multirow{2}{*}{Imb.} 
      & 2    
      & Bias       & 5  & \textbf{11} & 6  & 6  & 5  & 7  \\ 
    \cline{3-10}
      & 
      &       
      & Std.\ Dev. & 6  & \textbf{10} & 7  & 5  & 9  & 3  \\ 
    \cline{3-10}
      & 
      & 10   
      & Bias       & 5  & 5  & \textbf{11} & 8  & 8  & 3  \\ 
    \cline{3-10}
      & 
      &       
      & Std.\ Dev. & 9  & 7  & 5  & \textbf{10} & 7  & 2  \\ 
    \hline
  \end{tabular}
  \label{tbl:3wins}
\end{table*}

\begin{figure}[tph]
    \begin{center}
        \includegraphics[width=0.90\columnwidth]{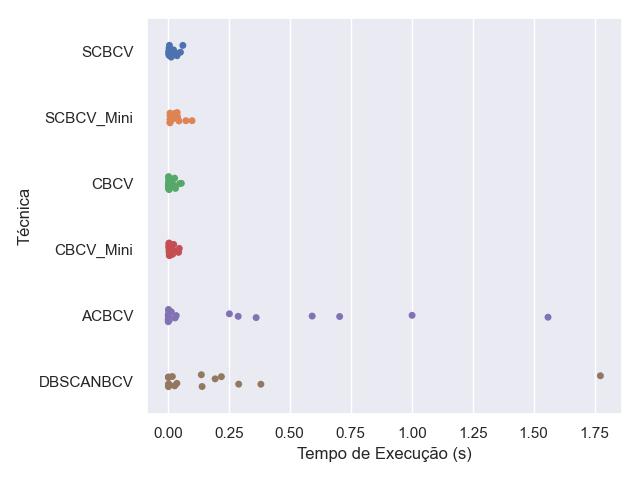}
    \end{center}
    \caption{Overall execution time of the techniques in Experiment Set 3 on the 20 datasets using 10 folds.}
    \label{fig:3runningtimes}
\end{figure}

Regarding computational cost, Figure \ref{fig:3runningtimes} reports execution times for each technique across the 20 datasets. The fastest methods were the KCBCV variants, followed by SCBCV, with DBSCANBCV and ACBCV being slower. As expected, the techniques based on K-Means, an intrinsically simpler clustering algorithm, ran more quickly than those relying on the more complex DBSCAN and Agglomerative Clustering algorithms.

\section{Conclusion}
\label{sec:conclusion}
The experiments conducted in this study enabled a detailed evaluation of various cross-validation techniques, considering bias, variance, and computational cost. While no single technique consistently outperformed the others across all scenarios, several important patterns emerged.

In balanced datasets, the technique proposed in this work, Stratified Cross-Validation using Mini-Batch K-Means (SCBCV Mini), demonstrated superior performance in terms of both bias and variance, maintaining values closer to zero when compared to Stratified Cross-Validation (SCV) and other clustering-based validation methods. This indicates that SCBCV Mini is effectively suited for scenarios where data distribution is balanced. However, the performance gains were not accompanied by a significant reduction in computational cost; in fact, in some cases, SCBCV Mini was slower than its counterpart using traditional K-Means (SCBCV), contrary to initial expectations.

In contrast, for imbalanced datasets, SCV consistently outperformed the other techniques. It exhibited lower bias and variance, especially in 10-fold settings, proving to be more robust and stable when dealing with uneven class distributions. Among the clustering-based techniques, KCBCV showed slight advantages, but its overall performance did not match that of SCV. In addition to being a well-established method, SCV also stood out for having the lowest computational cost, making it the preferred option in scenarios where class imbalance is a critical factor in model evaluation.
Although some of the cluster-based techniques, such as ACBCV and KCBCV, showed potential in certain contexts, occasionally with one technique outperforming another, their benefits were subtle. This suggests that the use of clustering algorithms in cross-validation must be carefully tailored to the structure and specific characteristics of the dataset under analysis. In the context of this study, which focused on evaluating the generalization ability of the techniques, no algorithm consistently emerged as superior. Therefore, the selection and configuration of clustering algorithms should be made with close attention to dataset-specific features to fully leverage the potential of cluster-based cross-validation.

In summary, the results indicate that while no cluster-based cross-validation technique clearly outperforms the others across all aspects, SCV remains the most balanced and efficient choice in class-imbalanced scenarios. The technique proposed in this study, SCBCV Mini, outperformed SCV and other clustering-based methods in balanced datasets, but at the cost of a slightly higher computational load. This unexpected cost may be attributed to the use of a fixed batch size of 1024 across all datasets, including those with few instances, a limitation that may have affected the performance of the algorithm.

For future work, it is recommended to investigate more efficient methods for determining the number of clusters in Agglomerative Clustering, considering different linkage types and the potential to optimize dendrogram cutting. Additionally, exploring the use of advanced hyperparameter optimization techniques, such as Hyperopt or Optuna, could yield significant improvements by efficiently exploring a wider range of configurations. These methods can enhance cross-validation by incorporating smarter search strategies and fine-tuning, leading to more robust and precise model evaluations.

Moreover, a more in-depth investigation into the computational complexity of each stage of the algorithm, particularly regarding the use of Mini-Batch K-Means, would be highly valuable. This should include an analysis of the relationship between batch size and algorithm performance across datasets of varying sizes, especially in Big Data scenarios. It is also advisable to conduct experiments with larger datasets that challenge the current methodological limitations, providing a stronger foundation for the application of robust techniques under high computational demand.

Additionally, incorporating cluster evaluation metrics, such as the silhouette score, and performing a more detailed analysis of the distribution between training and validation sets, comparing clustered versus non-clustered partitions, may offer further insights into the effectiveness of the proposed cross-validation strategies.

\bibliographystyle{IEEEtran}
\bibliography{biblio_ieee_ordered}

\end{document}